\documentclass{article}
\PassOptionsToPackage{numbers, compress}{natbib}
\usepackage[final]{neurips_2026}
\makeatletter
\providecommand{\@trackname}{}
\makeatother

\usepackage[utf8]{inputenc}
\usepackage[T1]{fontenc}
\usepackage{hyperref}
\usepackage{url}
\usepackage{booktabs}
\usepackage{amsfonts}
\usepackage{amsmath}
\usepackage{amssymb}
\usepackage{nicefrac}
\usepackage{multirow}
\usepackage{microtype}
\usepackage{xcolor}
\usepackage{tcolorbox}
\usepackage{graphicx}
\usepackage{float}
\usepackage{placeins}
\usepackage{subcaption}
\usepackage{wrapfig}
\usepackage{xspace}
\newcommand{\method}{{QViD}\xspace}

\definecolor{keyquestionrule}{gray}{0.72}
\definecolor{keyquestionfill}{gray}{0.985}
\newcommand{\keyquestion}[1]{%
  \begin{center}
  \begin{tcolorbox}[
    colback=keyquestionfill,
    colframe=keyquestionrule,
    boxrule=0.35pt,
    arc=3pt,
    outer arc=3pt,
    left=6pt,
    right=6pt,
    top=3pt,
    bottom=3pt,
    width=0.98\linewidth,
    before skip=0pt,
    after skip=0pt
  ]
  \centering\emph{#1}\par
  \end{tcolorbox}
  \end{center}
}
\title{
Vision Token Pruning via Query--Vision Interaction Decomposition}

\author{%
Harshithanjani Athi\textsuperscript{1}\thanks{%
Email addresses:
\texttt{ha26227.my@utexas.edu},
\texttt{sravan.ankireddy@utexas.edu},
\texttt{jianzhong.z@samsung.com},
\texttt{hyeji.kim@utexas.edu}.},
Sravan Kumar Ankireddy\textsuperscript{1},
Jianzhong Charlie Zhang\textsuperscript{2},
Hyeji Kim\textsuperscript{1}
\\[0.3em]
{\small
\textsuperscript{1}The University of Texas at Austin
\quad
\textsuperscript{2}Samsung Research America
}
}
\begin{document}

\maketitle
\begin{abstract}
Large vision-language models (VLMs), which process both visual inputs and text queries, incur high inference costs due to the large number of visual tokens, making visual token pruning a natural approach for improving efficiency. 
Recent work has leveraged query information to guide token selection, often through heuristics derived from attention patterns or token dynamics. 
However, these approaches do not explicitly exploit the underlying structure of query--vision interactions.
We address this gap by observing that the query--vision interaction matrix, naturally induced by the query and key projections already present in transformer attention, is effectively low-rank, with a small number of dominant latent interaction modes capturing query-relevant semantics.
Motivated by this observation, we propose Query-Vision Decomposition (\method), a novel training-free query-aware visual token pruning method that exploits this structure.
Extensive experiments on both image and video understanding benchmarks show that \method consistently improves performance under matched token budgets, with the largest gains appearing under aggressive compression regimes. 
\end{abstract}

\section{Introduction}
\label{sec:intro}

Vision-language models (VLMs) have demonstrated strong performance across a wide range of multimodal tasks, including visual question answering, optical character recognition, diagram understanding, and video comprehension~\cite{liu2024llava, wang2024qwen2vl, li2024llavaonevision}. In a typical VLM, a vision encoder produces vision tokens from an image, and the language model jointly processes these vision tokens together with the text query to generate a final response. However, this design also creates a major efficiency bottleneck: an image can produce hundreds to thousands of vision tokens, making multimodal inference expensive in both latency and memory. Consequently, reducing the cost of processing vision tokens has become an important focus of recent work on efficient VLM inference. 

A direct remedy is to reduce the number of visual tokens passed to the language model. Several \emph{training-free} token compression methods have been proposed for this purpose, and many of them use \emph{attention-guided} signals. For example, FastV~\cite{chen2024fastv} and SparseVLM~\cite{zhang2025sparsevlm} use early-layer attention patterns to score visual tokens, then prune tokens with low importance scores. While intuitive and often effective, these methods can be biased: vision tokens that are closer to the text tokens in the sequence may receive higher attention scores even when they are not the most relevant to the query. As a result, attention-based scores do not always provide a reliable measure of true query relevance. 

The second paradigm is \emph{attention-free} token pruning. A strong representative of this line of work is V$^2$Drop~\cite{chen2024v2drop}, which shows that an attention-free pruning signal can outperform earlier attention-based methods. Instead of using attention scores, it ranks vision tokens by how much their hidden states change across consecutive LLM layers, and drops tokens that change the least. This avoids the positional bias of attention-based scores. Since visual and text tokens are processed jointly, the variation signal can still reflect some influence from the query, but only indirectly. It captures generic token activity rather than explicitly measuring query--vision relevance.

\begin{figure*}[t]
    \centering
    \includegraphics[width=\textwidth]{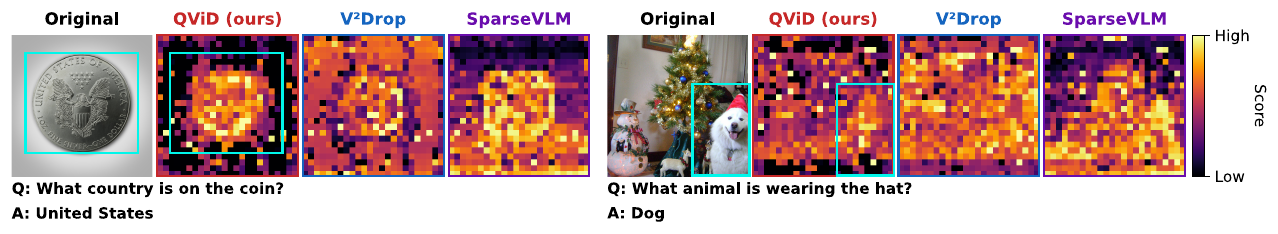}
  \caption{\textbf{Comparison of Visual-Token Scoring.} Compared with prior methods such as V$^2$Drop and SparseVLM, \method concentrates high scores on query-relevant image regions, leading to more effective visual-token pruning.}
    \label{fig:main_qualitative}
\end{figure*}

This reveals a gap between the two paradigms: attention-guided methods are query-aware but inherit attention-score biases, while variation-based methods avoid attention scores but do not explicitly model the query. Our goal is to combine the strengths of both, enabling a query-conditioned pruning signal that does not rely on attention weights. This is particularly important because, in visual question answering and related tasks, the query determines what is relevant in the image. 
For example, in an image of an animal wearing a hat, as in Figure~\ref{fig:main_qualitative}, tokens depicting the animal are highly relevant for identifying it, but less relevant for questions about the background or surrounding objects. Ideally, a token selection method should therefore explicitly condition on the query, and more precisely on its underlying semantics. This raises the question of how such semantics can be represented in a structured form. This motivates our central question:



\keyquestion{Can a few dominant latent modes of query--vision interaction capture query-relevant visual evidence without relying on attention weights?}

\paragraph{Our Approach.}
To answer this question, we propose \textbf{Q}uery-\textbf{Vi}sion \textbf{D}ecomposition (\method), a training-free method for visual-token selection based on an explicit query--vision interaction matrix. 
Our key observation is that query--vision interactions exhibit a strong low-rank structure: a small number of dominant interaction modes capture the visual patterns most relevant to the query.

Motivated by this observation, \method constructs a query--vision interaction matrix at an early transformer layer and scores visual tokens according to how strongly they align with these dominant modes. 
Unlike attention-guided pruning, \method does not rely on attention weights; unlike variation-based pruning, it explicitly conditions on the input query.

Figure~\ref{fig:main_qualitative} qualitatively compares \method with representative attention-guided (SparseVLM) and attention-free (V$^2$Drop) approaches. While these prior methods can highlight regions not directly relevant to the query, \method focuses on image regions that provide direct evidence for the answer. 

Our main contributions are summarized as follows:

\begin{itemize}
    \item We show that query--vision interactions in VLMs, readily obtainable from standard transformer query--key projections, exhibit a strong low-rank structure, where task-relevant semantics concentrate into a small number of dominant interaction modes.

    \item We propose \method, an SVD-based query-aware visual-token pruning method that estimates token importance from the dominant singular structure of the query--vision interaction matrix. 
    Our analysis shows that isolating these dominant interaction modes is crucial for effective visual-token selection.

    \item Experiments across multiple VLMs and image and video understanding benchmarks show that \method consistently outperforms strong recent pruning methods across a wide range of token budgets, with particularly strong gains under aggressive compression.
\end{itemize}

\section{Related Work}
\label{sec:related}


Modern vision-language models (VLMs) process visual and textual tokens jointly, and recent advances toward higher-resolution images and longer videos have significantly increased the number of visual tokens~\cite{liu2023llava,liu2024llava,wang2024qwen2vl,li2024llavaonevision,lin2023videollava}. 
This has led to a growing line of work on visual-token compression for efficient VLM inference. 
In this section, we review existing approaches to visual-token compression and position our method within this landscape. Our focus is on {\em training-free visual-token pruning}, which is complementary to improvements in the vision encoder, projector, or LLM backbone.


\paragraph{Visual-Token Compression.}
A common approach to reducing visual-token cost is to prune or merge tokens based on importance scores. Attention-guided methods such as FastV~\cite{chen2024fastv} and SparseVLM~\cite{zhang2025sparsevlm} use attention patterns inside the LLM to identify important visual tokens, while PDrop~\cite{xing2025pdrop} progressively reduces the token count across layers. Token merging methods such as ToMe~\cite{bolya2023tome} and LLaVA-PruMerge~\cite{shang2025llavaprumerge} reduce sequence length by combining similar visual tokens instead of discarding them.  While effective, attention-based scoring requires access to attention matrices, making it less compatible with memory-efficient implementations such as FlashAttention~\cite{dao2024flashattention}, and can suffer from positional bias in autoregressive LLMs~\cite{chen2024v2drop}.


\paragraph{Task-Agnostic Versus Query-Aware Pruning.}
V$^2$Drop~\cite{chen2024v2drop} avoids explicit attention scores by measuring how much each visual token changes across consecutive LLM layers. Tokens with small hidden-state variation are treated as ``lazy tokens'' and pruned. This makes V$^2$Drop naturally compatible with FlashAttention and avoids attention-position bias. However, the variation signal is largely task-agnostic: it captures generic token dynamics rather than explicitly modeling query--vision relevance. 
In contrast, \method constructs a query--vision interaction matrix and ranks tokens based on their participation in its dominant low-rank structure, yielding a training-free pruning rule that is explicitly conditioned on the input query.

\begin{figure}[t]
    \centering
    \includegraphics[width=\linewidth]{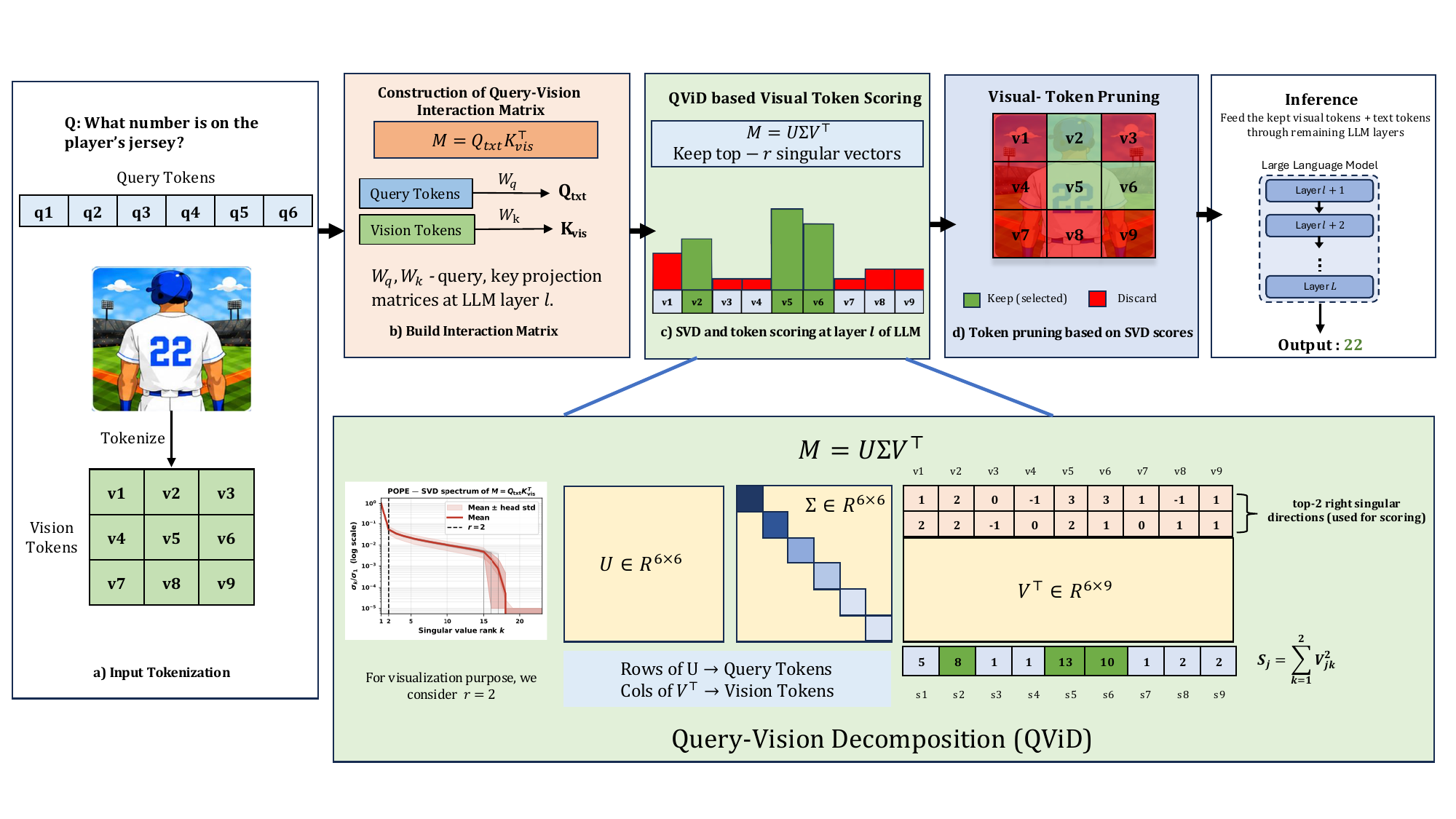}
    \caption{\textbf{\method Pipeline.} \method performs query-aware visual-token pruning by explicitly modeling query--vision interactions. At pruning layer $\ell$, it forms the interaction matrix $M = Q_{\text{txt}}K_{\text{vis}}^\top$, extracts its dominant low-rank modes with SVD, and scores each visual token by its participation in these modes. The top-$K$ tokens are retained for subsequent reasoning, preserving query-relevant visual evidence while reducing sequence length. Integer values in the illustrative matrices are used only for ease of presentation.}
    
    \label{fig:qvissvd_pipeline}
\end{figure}

\section{Method}
\label{sec:method}


\subsection{Problem Setup}
We consider a vision-language model (VLM) that processes an input image and a text query jointly. 
The image is first encoded into a sequence of visual tokens, which are then combined with text tokens and processed by a transformer-based language model. At transformer layer \(\ell\), let
\[
H^{(\ell)}_{\mathrm{vis}} \in \mathbb{R}^{N_v \times d},
\qquad
H^{(\ell)}_{\mathrm{txt}} \in \mathbb{R}^{N_q \times d}
\]
denote the hidden states corresponding to the visual tokens and user-query tokens, respectively, where \(N_v\) and \(N_q\) are the numbers of visual and query tokens, and \(d\) is the hidden dimension.

Our goal is to prune {\em visual tokens} in a query-aware and training-free manner by retaining those most relevant to the input query under a target token budget.  
Our method can be applied at any layer \(\ell\), and we study this design choice empirically.


\subsection{Query--Vision Interaction Matrix}
\label{sec:method:qvim}

Our method is based on a query--vision interaction matrix that measures the compatibility between query tokens and visual tokens. 
We construct this matrix using the frozen query and key projections already present in the transformer. For a given attention head at layer \(\ell\), we compute
\begin{equation}
    Q_{\mathrm{txt}} = H^{(\ell)}_{\mathrm{txt}} W_Q,
    \qquad
    K_{\mathrm{vis}} = H^{(\ell)}_{\mathrm{vis}} W_K,
\end{equation}
where \(W_Q, W_K \in \mathbb{R}^{d \times d_h}\) are the query and key projection matrices, and \(d_h\) is the attention-head dimension. We then define the query--vision interaction matrix
\begin{equation}
    M = Q_{\mathrm{txt}} K_{\mathrm{vis}}^\top
    \in \mathbb{R}^{N_q \times N_v}.
    \label{eq:qvim}
\end{equation}

The entry \(M_{ij}\) measures the compatibility between query token \(i\) and visual token \(j\) in the model's learned representation space. 
Equivalently, the \(j\)-th column of \(M\) summarizes how visual token \(j\) interacts with the full query.

\textbf{Advantages over Attention Scores.} This construction is closely related to the pre-softmax attention computation, but differs in two important ways. First, we compute the interaction before positional encoding. In autoregressive LLMs, positional encoding can bias attention toward visual tokens that are closer to the query tokens in the sequence, regardless of their semantic relevance. Since several prior pruning methods, such as SparseVLM and FastV, rely on attention scores for token selection, this positional bias can directly affect pruning decisions. By removing positional effects, \(M\) captures a purely content-based query--vision interaction signal.

Second, we avoid the normalization effects of softmax, where token relevance is mixed with competition among tokens. As a result, \(M\) preserves the raw query--vision interaction structure without relying on explicit attention maps. This is also advantageous in memory-efficient attention implementations, where attention maps may be unavailable or costly to extract.

\textbf{From Interaction Matrix to Token Scores.}
A natural way to derive visual-token importance from \(M\) would be to aggregate each column independently, for example by taking its norm. 
This measures how strongly a visual token is compatible with all query tokens. 
However, as we show later in the ablation study, naive column-wise aggregation performs poorly as a pruning criterion.

This is because not every query token, or word, is equally informative or relevant for the downstream task. 
For example, in the question ``What animal is wearing the hat?'', we would like to emphasize compatibility with meaningful words such as {\em animal} and {\em hat}, rather than less informative words such as {\em is} or {\em the}. 

More broadly, our goal is to capture visual evidence relevant to the {\em overall query semantics}, rather than simply measuring isolated token-level compatibility. 
As such relevance may not be recoverable from pairwise query--vision interactions alone, we instead analyze the global structure of the interaction matrix \(M\) itself.
\subsection{Low-Rank Structure of Query--Vision Interactions and Dominant Interaction Modes}
\label{sec:method:motivation}

Interestingly, we observe that \(M\), a quantity closely related to attention computation but not typically analyzed directly in prior pruning methods, exhibits a strong low-rank structure across a wide range of queries and images. 
Moreover, the dominant singular modes are not merely numerically significant; they consistently capture task-relevant query--vision interaction patterns.

\begin{figure}[t]
    \centering
    \includegraphics[width=0.30\linewidth]{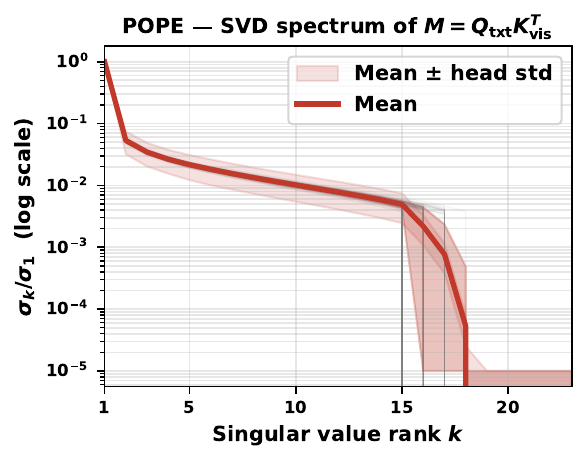}
    \hfill
    \includegraphics[width=0.68\linewidth]{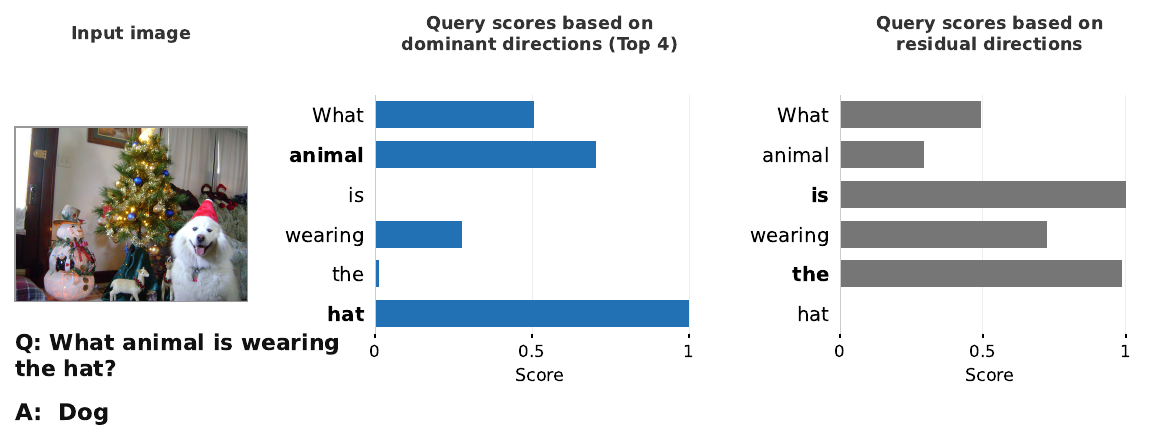}
    \caption{\textbf{Qualitative Observations.}
    \textbf{(a)} The query--vision interaction matrix has a rapidly decaying singular-value spectrum. 
    \textbf{(b)} Dominant SVD components emphasize query-relevant tokens, while residual components are less aligned with the task.}
    \label{fig:obs_side_by_side}
\end{figure}
\paragraph{Observation 1: Query--Vision Interactions Are Approximately Low-Rank.}
Across images and queries, the singular spectrum of \(M\) is strongly concentrated in a small number of directions. 
For each attention head, we compute the singular value decomposition (SVD) \(M = U \Sigma V^\top\). 

Figure~\ref{fig:obs_side_by_side} (a) shows the singular value spectrum of \(M\) for LLaVA-1.5-7B on the POPE dataset~\cite{li2023pope}. 
We observe that the spectrum decays rapidly, with a small number of singular values explaining most of the query--vision interaction structure. 
As further shown in Appendix~\ref{app:analysis_spectrum}, this behavior appears consistently across different datasets, images, and queries, suggesting that the useful interaction between query and visual tokens often lies in a low-dimensional subspace.

\paragraph{Observation 2: Dominant Components Capture Task-Relevant Structure.}
The low-rank structure of \(M\) is not only a numerical property; it is also semantically meaningful. In particular, we observe that task-relevant semantics tend to concentrate into a small number of dominant interaction modes.

For interpretability, we separate the leading singular directions, which capture the dominant interaction modes, from the remaining residual directions. 
We then measure how strongly each query token $j$ participates in the dominant and residual subspaces through
\begin{equation}
s_j^{(\mathrm{dom})} = \sum_{k=1}^{4} U_{jk}^2,
\qquad
s_j^{(\mathrm{res})} = \sum_{k>4} U_{jk}^2.
\end{equation}
Figure~\ref{fig:obs_side_by_side} (b) visualizes these two scores for the dog query--image example; additional qualitative examples are provided in Appendix~\ref{app:qualitative_svd_components}.  
We observe that semantically informative query tokens, such as \emph{animal} and \emph{hat}, tend to have high dominant contribution scores and low residual contribution scores. 
In contrast, less informative words, such as \emph{is} and \emph{the}, exhibit the opposite behavior.

Together, these observations suggest the following: 

\keyquestion{Visual-token importance should be estimated from the dominant interaction subspace of \(M\), rather than from individual matrix entries or raw column magnitudes.}

\subsection{\method Token Scoring and Pruning}
\label{sec:method:qvissvd}

We now describe how \method  converts the query--vision interaction matrix into visual-token scores. 

{\bf Visual-Token Scores.}
For each attention head at layer \(\ell\), we compute the singular value decomposition
\begin{equation}
    M = U \Sigma V^\top .
\end{equation}
The right singular vectors in \(V\) describe the visual-token side of the query--vision interaction modes. 
We score each visual token \(j\) by its participation in the top \(r\) right singular directions:
\begin{equation}
    s_j = \sum_{k=1}^{r} V_{jk}^{2}.
    \label{eq:qvissvd_score}
\end{equation}

Intuitively, \(V_{jk}^2\) measures how strongly visual token \(j\) participates in the \(k\)-th dominant interaction mode. Therefore, a larger \(s_j\) indicates that the visual token \(j\) contributes more strongly to the dominant interaction subspace. 

For the multi-head setting, we apply the same scoring rule independently to each attention head and average the resulting scores. 
Let \(M^{(h)}\) denote the query--vision interaction matrix for head \(h\), and let \(V^{(h)}\) be its right singular vector matrix. 
The final score for visual token \(j\) is
\begin{equation}
    s_j =
    \frac{1}{H}
    \sum_{h=1}^{H}
    \sum_{k=1}^{r}
    \left(V^{(h)}_{jk}\right)^2,
    \label{eq:multi_head_score}
\end{equation}
where \(H\) is the number of attention heads. 

For short queries, the number of available nonzero singular directions may be smaller than the chosen \(r\). 
In this case, we use $r_{\mathrm{eff}} = \min(r, N_q, N_v)$, 
where \(N_q\) and \(N_v\) denote the numbers of query and visual tokens, respectively, and replace \(r\) by \(r_{\mathrm{eff}}\) in the scoring rule.

{\bf Visual-Token Pruning.} The top-\(K\) visual tokens with the largest scores \(s_j\) are passed to the remaining LLM layers, with \(K\) determined by the token budget.
Importantly, \method is entirely training-free, requires no architectural modification, and operates solely using the frozen projections already present in the model. 

\section{Experiments}
\label{sec:experiments}


\subsection{Experimental Setup}

We conduct comprehensive experiments across image understanding, video understanding, multimodal reasoning, hallucination-sensitive evaluation, and captioning tasks under a wide range of token budgets. In the main paper, we focus on the core image and video understanding results, while Appendices~\ref{app:additional_multimodal_benchmarks},\ref{app:hallusion_coco} further demonstrate the robustness and generality of \method across diverse settings. 
Specifically, we consider the following models, benchmarks, and baselines.

\textbf{Models.}
We evaluate \method on four representative VLMs. 
For image understanding, we use LLaVA-1.5-7B~\cite{liu2024llava} and Qwen2-VL-7B~\cite{wang2024qwen2vl}. 
For video understanding, we use Video-LLaVA-7B~\cite{lin2023videollava} and LLaVA-OneVision-7B~\cite{li2024llavaonevision}.

\textbf{Benchmarks.}
For image understanding, we consider standard and diverse VLM benchmarks, including GQA~\cite{hudson2019gqa}, ScienceQA~\cite{lu2022scienceqa}, TextVQA~\cite{singh2019textvqa}, POPE~\cite{li2023pope}, MME~\cite{fu2023mme}, and MMBench~\cite{liu2024mmbench}. For video understanding, we evaluate on VideoMME~\cite{fu2024videomme} and MVBench~\cite{li2024mvbench}. 


\textbf{Baselines.}
V$^2$Drop~\cite{chen2024v2drop} serves as the primary baseline throughout our experiments, as it is the most recent and strongest existing visual-token pruning method. For LLaVA-1.5-7B, we additionally compare against ToMe~\cite{bolya2023tome}, FastV~\cite{chen2024fastv}, HiRED~\cite{arif2025hired}, LLaVA-PruMerge~\cite{shang2025llavaprumerge}, SparseVLM~\cite{zhang2025sparsevlm}, and PDrop~\cite{xing2025pdrop}. For these methods, we use the numbers reported in the V$^2$Drop paper. For V$^2$Drop itself, we reproduce the results using the official implementation and confirm that they match the reported numbers.

\textbf{Implementation Details.}
All experiments are conducted using Python 3.11, PyTorch 2.5.1 (CUDA 12.1). For LLaVA-1.5-7B, Qwen2-VL-7B, Video-LLaVA-7B, and LLaVA-OV-7B, we follow the standard evaluation settings of the corresponding model and benchmark. \method is implemented as a separate scoring module at the pruning layer and does not modify the model architecture or attention kernel. After token selection, all subsequent transformer layers process the retained sequence normally. We use layer \(\ell=3\), \(r=16\) SVD components, and omit Rotary Positional Embedding (RoPE) when constructing the query--vision interaction matrix. 



\subsection{\method: Main Results}\label{sec:results}

Across all evaluated VLMs and benchmarks, \method consistently achieves stronger performance under matched token budgets. The gains are particularly pronounced under aggressive compression regimes, where retaining only a small fraction of visual tokens makes query-aware selection especially important.

Compared with existing pruning methods, \method more effectively preserves task-relevant visual information, leading to improved downstream accuracy on both image and video understanding tasks. 
This behavior is also reflected {\em qualitatively} in the retained visual tokens and score maps, as illustrated in Figure~\ref{fig:main_qualitative} and additional visualizations provided in Appendix~\ref{app:additional_visualizations}.

\paragraph{LLaVA-1.5-7B Image Understanding.} Table~\ref{tab:llava} compares \method with prior visual-token compression methods under matched token budgets.
We report the mean normalized performance relative to the full-token baseline. \method achieves the best average performance at all three token budgets. 
The gains are largest at 64 tokens, where \method improves over V$^2$Drop by \textbf{3.5}$\%$ in normalized average, with particularly strong gains on TextVQA (3.7$\%$), POPE (5.7$\%$), and MME (83 points). This suggests that explicit query conditioning is especially important under aggressive compression, where query-agnostic signals may retain irrelevant visual tokens.

\begin{table}[tb]
\caption{\textbf{Image Understanding Results on LLaVA-1.5-7B.}
\method achieves the best average performance at all retention ratios, with the largest gain under aggressive compression at 64 tokens. 
\textit{Methods are sorted by average performance within each retention setting.}}
\label{tab:llava}
\centering
\small
\setlength{\tabcolsep}{4pt}
\renewcommand{\arraystretch}{0.98}
\begin{tabular}{lccccccc}
\toprule
Methods & GQA & SQA & TextVQA & POPE & MME & MMB & Average \\
\midrule
\multicolumn{8}{l}{\textit{Upper Bound, 576 Tokens}} \\
LLaVA-1.5-7B (Liu et al. 2024) & 61.9 & 69.5 & 58.2 & 85.9 & 1862 & 64.6 & 100.0\% \\
\midrule

\multicolumn{8}{c}{\textit{Average Retain 192 Tokens ($\downarrow$66.7\%)}} \\
\midrule
\textbf{\method (ours)} & \textbf{59.3} & 69.0 & \textbf{57.8} & \textbf{87.4} & 1793 & 63.2 & \textbf{98.1\%} \\
\addlinespace[1pt]
V$^2$Drop~[CVPR'26]    & 58.5 & \textbf{69.3} & 55.6 & 85.1 & \textbf{1826} & \textbf{63.7} & 97.6\% \\
PDrop~[CVPR'25]        & 57.1 & 68.8 & 56.1 & 82.3 & 1766 & 63.2 & 96.0\% \\
SparseVLM~[ICML'25]    & 57.6 & 69.1 & 56.1 & 83.6 & 1721 & 62.5 & 95.9\% \\
HiRED~[AAAI'25]        & 58.7 & 68.4 & 47.4 & 82.8 & 1737 & 62.8 & 93.6\% \\
LLaVA-PruMerge~[ICCV'25] & 54.3 & 67.9 & 54.3 & 71.3 & 1632 & 59.6 & 90.3\% \\
ToMe~[ICLR'23]         & 54.3 & 65.2 & 52.1 & 72.4 & 1563 & 60.5 & 88.8\% \\
FastV~[ECCV'24]        & 52.7 & 67.3 & 52.5 & 64.8 & 1612 & 61.2 & 88.2\% \\
\midrule

\multicolumn{8}{c}{\textit{Average Retain 128 Tokens ($\downarrow$77.8\%)}} \\
\midrule
\textbf{\method (ours)} & \textbf{57.7} & 68.7 & \textbf{57.1} & \textbf{85.5} & \textbf{1726} & \textbf{61.9} & \textbf{96.4\%} \\
\addlinespace[1pt]
V$^2$Drop~[CVPR'26]    & 56.3 & \textbf{68.8} & 53.8 & 80.9 & 1712 & 61.8 & 94.0\% \\
PDrop~[CVPR'25]        & 56.0 & 68.3 & 54.8 & 82.3 & 1644 & 61.1 & 93.6\% \\
SparseVLM~[ICML'25]    & 56.0 & 67.1 & 54.9 & 80.5 & 1696 & 60.0 & 93.2\% \\
HiRED~[AAAI'25]        & 57.2 & 68.1 & 46.1 & 79.8 & 1710 & 61.5 & 91.6\% \\
LLaVA-PruMerge~[ICCV'25] & 53.3 & 67.1 & 54.3 & 67.2 & 1554 & 58.1 & 87.9\% \\
FastV~[ECCV'24]        & 49.6 & 60.2 & 50.6 & 59.6 & 1490 & 56.1 & 81.7\% \\
ToMe~[ICLR'23]         & 52.4 & 59.6 & 49.1 & 62.8 & 1343 & 53.3 & 80.4\% \\
\midrule

\multicolumn{8}{c}{\textit{Average Retain 64 Tokens ($\downarrow$88.9\%)}} \\
\midrule
\textbf{\method (ours)} & 52.2 & 68.3 & \textbf{55.5} & \textbf{80.8} & \textbf{1553} & \textbf{56.3} & \textbf{90.4\%} \\
\addlinespace[1pt]
V$^2$Drop~[CVPR'26]    & 50.5 & \textbf{68.9} & 51.8 & 75.1 & 1470 & 55.2 & 86.9\% \\
SparseVLM~[ICML'25]    & \textbf{52.7} & 62.2 & 51.8 & 75.1 & 1505 & 56.2 & 86.5\% \\
LLaVA-PruMerge~[ICCV'25] & 51.9 & 68.1 & 54.0 & 65.3 & 1549 & 55.2 & 86.5\% \\
PDrop~[CVPR'25]        & 41.9 & 68.6 & 45.9 & 55.9 & 1092 & 33.3 & 70.1\% \\
FastV~[ECCV'24]        & 46.1 & 51.1 & 47.8 & 48.0 & 1256 & 48.0 & 71.3\% \\
ToMe~[ICLR'23]         & 48.6 & 50.0 & 45.3 & 52.5 & 1138 & 43.7 & 69.7\% \\
\bottomrule
\end{tabular}
\vspace{-1em}
\end{table}

\paragraph{Qwen2-VL-7B Image Understanding.}
Table~\ref{tab:qwen} shows that the gains of \method generalize to Qwen2-VL-7B. 
Across all retention ratios, \method improves the average normalized performance over V$^2$Drop. 
The improvement is particularly large at 11.1\% retention, where \method improves the average from 62.7\% to 81.7\%, indicating that query-aware scoring is especially important when only a small fraction of visual tokens can be retained.

\begin{table*}[!htb]
\caption{\textbf{Image Understanding Results on Qwen2-VL-7B.}
\method improves average performance at all retention ratios, with the largest gains at 11.1\% retention. 
Avg. is mean normalized performance relative to the full-token baseline.}
\label{tab:qwen} 
\centering
\footnotesize
\setlength{\tabcolsep}{3.6pt}
\renewcommand{\arraystretch}{0.96}
\begin{tabular}{@{}lccccccccc@{}}
\toprule
Method & AI2D & GQA & TextVQA & POPE & MME-P & MME-C & SQA & MMB & Avg. \\
\midrule
Baseline & 80.7 & 61.9 & 81.5 & 88.2 & 1684.0 & 635.7 & 85.3 & 84.9 & 100.0\% \\
\midrule

\multicolumn{10}{c}{\textit{Retain 33.3\%} ($\downarrow$66.7\%)} \\
\midrule
\textbf{\method (ours)} & 76.6 & \textbf{58.6} & \textbf{79.0} & \textbf{85.6} & \textbf{1610.6} & 493.6 & \textbf{81.2} & 81.0 & \textbf{93.4\%} \\
V$^2$Drop~[CVPR'26]  & \textbf{76.7} & 52.8 & 53.4 & 85.4 & 1589.0 & \textbf{600.4} & 80.9 & \textbf{82.0} & 90.4\% \\
\midrule

\multicolumn{10}{c}{\textit{Retain 22.2\%} ($\downarrow$77.8\%)} \\
\midrule
\textbf{\method (ours)} & \textbf{73.8} & \textbf{56.0} & \textbf{77.1} & \textbf{83.7} & \textbf{1577.9} & 446.1 & \textbf{79.0} & \textbf{78.8} & \textbf{90.1\%} \\
V$^2$Drop~[CVPR'26]  & 71.9 & 49.6 & 52.1 & 81.2 & 1563.3 & \textbf{482.9} & 78.4 & 78.6 & 84.8\% \\
\midrule

\multicolumn{10}{c}{\textit{Retain 11.1\%} ($\downarrow$88.9\%)} \\
\midrule
\textbf{\method (ours)} & \textbf{69.8} & \textbf{50.9} & \textbf{69.1} & \textbf{78.4} & \textbf{1420.1} & \textbf{352.5} & \textbf{75.3} & \textbf{70.7} & \textbf{81.7\%} \\
V$^2$Drop~[CVPR'26]  & 65.0 & 39.7 & 45.0 & 57.6 & 863.0 & 294.6 & 70.1 & 48.4 & 62.7\% \\
\bottomrule
\end{tabular}
\end{table*}
\paragraph{Video Understanding.}
Table~\ref{tab:video_and_compute}(a) shows that \method also extends effectively to video understanding on LLaVA-OV-7B.
\method consistently outperforms V$^2$Drop under matched token budgets, with a particularly large gain at 10\% retention, where it preserves 91.2\% of full-token performance compared to 69.4\% for V$^2$Drop. Additional results on Video-LLaVA-7B are provided in Appendix~\ref{app:video_llava_results}.

\paragraph{Additional Results.} Beyond image and video understanding, we also evaluate whether the selected tokens preserve broader multimodal capabilities.
Appendix~\ref{app:additional_multimodal_benchmarks} reports additional results on multimodal reasoning benchmarks, while Appendix~\ref{app:hallusion_coco} evaluates hallucination-sensitive performance and captioning quality. These results show that \method preserves strong performance beyond standard perception tasks, suggesting that its selected tokens retain the visual information needed for downstream reasoning and generation.

\begin{table*}[t]
\caption{\textbf{Video Understanding and Compute Comparison.}
(a) Results on LLaVA-OV-7B show that \method consistently outperforms V$^2$Drop under matched token budgets, with the largest gains under aggressive compression. 
Avg. denotes the mean normalized performance relative to the full-token baseline. 
(b) Total TFLOPs on LLaVA-1.5-7B include pruning overhead; reduction is computed relative to the full-token baseline.}
\centering
\footnotesize
\renewcommand{\arraystretch}{1.05}

\begin{minipage}[t]{0.66\textwidth}
\centering
\textbf{\strut (a) Video Understanding on LLaVA-OV-7B}\par\vspace{0.35em}
\setlength{\tabcolsep}{3pt}
\begin{tabular*}{\linewidth}{@{\extracolsep{\fill}}lcccccc@{}}
\toprule
\textbf{Method} & \textbf{MVBench} & \multicolumn{4}{c}{\textbf{VideoMME}} & \textbf{Avg.} \\
\cmidrule(lr){3-6}
& & \textbf{Overall} & \textbf{Short} & \textbf{Med.} & \textbf{Long} & \\
\midrule
Baseline & 58.4 & 58.2 & 70.3 & 56.4 & 47.9 & 100.0\% \\
\midrule

\multicolumn{7}{@{}c@{}}{\textit{Retention = 25\%}} \\
\method & \textbf{56.6} & \textbf{57.2} & \textbf{68.3} & \textbf{54.4} & 49.2 & \textbf{97.7\%} \\
V$^2$Drop~[CVPR'26] & 56.5 & 57.1 & 68.1 & 53.6 & \textbf{49.8} & 97.5\% \\
\midrule

\multicolumn{7}{@{}c@{}}{\textit{Retention = 15\%}} \\
\method & \textbf{54.0} & \textbf{55.7} & \textbf{65.0} & \textbf{53.1} & \textbf{49.1} & \textbf{94.2\%} \\
V$^2$Drop~[CVPR'26] & 53.0 & 53.7 & 62.3 & 51.6 & 47.1 & 91.7\% \\
\midrule

\multicolumn{7}{@{}c@{}}{\textit{Retention = 10\%}} \\
\method & \textbf{52.5} & \textbf{53.7} & \textbf{60.8} & \textbf{52.9} & \textbf{47.3} & \textbf{91.2\%} \\
V$^2$Drop~[CVPR'26] & 39.4 & 41.4 & 41.0 & 42.6 & 40.6 & 69.4\% \\
\bottomrule
\end{tabular*}
\end{minipage}
\hfill
\begin{minipage}[t]{0.31\textwidth}
\centering
\textbf{\strut (b) Compute Comparison}\par\vspace{0.35em}
\begingroup
\setlength{\tabcolsep}{3pt}
\renewcommand{\arraystretch}{1.05}
\begin{tabular*}{\linewidth}{@{\extracolsep{\fill}}lccc@{}}
\toprule
\textbf{Method} & \multicolumn{2}{c}{\textbf{Compute}} & \textbf{Red.} \\
\cmidrule(lr){2-3}
& \textbf{Tok.} & \textbf{TFLOPs} & \\
\midrule
Baseline & 576 & 8.86 & -- \\
\midrule
\multicolumn{4}{@{}c@{}}{\textit{Tokens = 64}} \\
V$^2$Drop & 64  & 1.97 & 77.7\% \\
\method  & 64  & 2.525 & 71.5\% \\
\midrule
\multicolumn{4}{@{}c@{}}{\textit{Tokens = 128}} \\
V$^2$Drop & 128 & 2.84 & 67.9\% \\
\method  & 128 & 3.421 & 61.4\% \\
\midrule
\multicolumn{4}{@{}c@{}}{\textit{Tokens = 192}} \\
V$^2$Drop & 192 & 3.72 & 57.9\% \\
\method  & 192 & 4.317 & 51.3\% \\
\bottomrule
\end{tabular*}
\endgroup
\end{minipage}

\vspace{0.5em}

\label{tab:video_and_compute}
\end{table*}

\textbf{Computational Complexity.}
Table~\ref{tab:video_and_compute} (b) reports total TFLOPs on LLaVA-1.5-7B, including pruning overhead: the full-token baseline costs \(8.86\) TFLOPs, while \method costs \(2.53\), \(3.42\), and \(4.32\) TFLOPs at \(64\), \(128\), and \(192\) retained tokens, reducing compute by \(71.5\%\), \(61.4\%\), and \(51.3\%\), respectively. Compared with V$^2$Drop, \method spends modest extra compute on query-aware scoring but retains more task-relevant visual tokens; detailed latency, throughput, FLOPs accounting, and accuracy--compute tradeoffs are provided in Appendix~\ref{app:efficiency}.

\subsection{Ablation Studies}
\label{sec:analysis}

We conduct ablation studies to examine the main design choices in \method.
In particular, we study whether the SVD-based decomposition is necessary, how many dominant SVD components are sufficient for token scoring, and how the choice of pruning layer affects performance.

\paragraph{Dominant Low-Rank Structure Is Important.}
To isolate the effect of SVD-based decomposition and low-rank interaction modeling, we compare \method with a simple column-norm baseline that uses the same query--vision interaction matrix \(M=Q_{\mathrm{txt}}K_{\mathrm{vis}}^\top\), but scores each visual token by raw interaction magnitude: $    s_j^{\mathrm{col}}=\|M_{:,j}\|_2^2$. 
This baseline directly measures how strongly each visual token interacts with the query, but does not separate dominant query-conditioned structure from residual interaction energy. 

Table~\ref{tab:scoring_ablation_main} (a) shows that simply using the query--vision interaction matrix is not sufficient. 
Column-norm scoring performs much worse than \method across all token budgets, especially under aggressive compression. 
The SVD step is crucial as it identifies the low-rank interaction modes that provide a more reliable signal for visual-token selection.

\paragraph{Effect of the Number of SVD Components.}
We study how many dominant SVD components are needed for reliable token scoring. Table~\ref{tab:scoring_ablation_main} (b) reports the component sweep at \(128\) retained tokens. We denote the number of SVD components used for scoring by \(n_{\mathrm{svd}}\).
A small number of dominant components is sufficient to recover most of the benefit of \method: performance improves sharply from \(n_{\mathrm{svd}}=1,2\) to \(n_{\mathrm{svd}}=4,8\), indicating that the most relevant query--vision structure is concentrated in a few dominant interaction modes.
Using \(n_{\mathrm{svd}}=16\) provides additional robustness and achieves the best average performance at this budget. 
The full sweep over all token budgets is provided in Appendix~\ref{app:nsvd_sweep}.

\begin{table*}[t]
\caption{\textbf{Ablations on \method Scoring with LLaVA-1.5-7B.}
(a) SVD scoring outperforms column-norm scoring from the same query--vision interaction matrix. 
(b) Using only a small number of dominant SVD components recovers most of the benefit of \method, indicating that token relevance is concentrated in a few interaction modes.}
\centering
\small
\setlength{\tabcolsep}{4pt}
\renewcommand{\arraystretch}{1.05}

\begin{minipage}[t]{0.49\textwidth}
\centering
\textbf{\strut (a) Effect of SVD-Based Scoring}\par\vspace{0.4em}
\resizebox{\linewidth}{!}{
\begin{tabular}{lcccccc}
\toprule
Method & GQA & SQA & TextVQA & POPE & MME & Avg. \\
\midrule
Baseline & 61.9 & 69.5 & 58.2 & 85.9 & 1862.0 & 100.0\% \\
\midrule
\multicolumn{7}{c}{\textit{Retain 64 Tokens ($\downarrow$88.9\%)}} \\
\midrule
\method   & \textbf{52.2} & \textbf{68.3} & \textbf{55.5} & \textbf{80.8} & \textbf{1553.0} & \textbf{91.1\%} \\
Column norm & 42.2 & 64.0 & 42.6 & 54.7 & 1009.0 & 70.3\% \\
\midrule
\multicolumn{7}{c}{\textit{Retain 128 Tokens ($\downarrow$77.8\%)}} \\
\midrule
\method   & \textbf{57.7} & \textbf{68.7} & \textbf{57.1} & \textbf{85.5} & \textbf{1726.0} & \textbf{96.5\%} \\
Column norm & 45.3 & 63.7 & 43.4 & 59.3 & 1069.0 & 73.2\% \\
\midrule
\multicolumn{7}{c}{\textit{Retain 192 Tokens ($\downarrow$66.7\%)}} \\
\midrule
\method   & \textbf{59.3} & \textbf{69.0} & \textbf{57.8} & \textbf{87.4} & \textbf{1793.0} & \textbf{98.5\%} \\
Column norm & 47.8 & 63.0 & 43.6 & 65.0 & 1190.0 & 76.5\% \\
\bottomrule
\end{tabular}
}
\end{minipage}
\hfill
\begin{minipage}[t]{0.485\textwidth}
\centering
\textbf{\strut (b) Number of SVD Components at 128 Tokens}\par\vspace{0.4em}
\begingroup
\renewcommand{\arraystretch}{1.7}
\resizebox{0.99\linewidth}{!}{
\begin{tabular}{lcccccc}
\toprule
Method & GQA & SQA & TextVQA & POPE & MME & Avg. \\
\midrule
Baseline & 61.9 & 69.5 & 58.2 & 85.9 & 1862.0 & 100.0\% \\
\midrule
V$^2$Drop              & 56.3 & \textbf{68.8} & 53.8 & 80.9 & 1712.0 & 93.7\% \\
\(n_{\mathrm{svd}}=1\)  & 45.6 & 63.4 & 43.5 & 35.5 & 1077.1 & 67.8\% \\
\(n_{\mathrm{svd}}=2\)  & 50.3 & 68.3 & 49.6 & 61.7 & 1478.6 & 83.2\% \\
\(n_{\mathrm{svd}}=4\)  & 56.7 & 68.5 & 56.0 & 80.1 & 1697.0 & 94.2\% \\
\(n_{\mathrm{svd}}=8\)  & 57.3 & 68.6 & 57.0 & 82.6 & 1725.1 & 95.6\% \\
\(n_{\mathrm{svd}}=16\) & \textbf{57.7} & 68.7 & \textbf{57.1} & \textbf{85.5} & \textbf{1726.0} & \textbf{96.5\%} \\
\bottomrule
\end{tabular}
}
\endgroup
\end{minipage}

\vspace{0.5em}
\label{tab:scoring_ablation_main}
\vspace{-1em}
\end{table*}

\paragraph{Choice of Pruning Layer and Other Design Choices.} Unless otherwise stated, we prune at LLM layer $\ell=3$, which provides a strong accuracy--efficiency tradeoff in our layer-selection study.
Detailed layer-selection results for both LLaVA-1.5-7B and Qwen2-VL-7B are provided in Appendix~\ref{app:layer_selection}. 
%
%
In addition, Appendix~\ref{app:scoring_rule_ablation} compares \method with alternative scoring rules derived from the same query--vision interaction matrix, including column-norm scoring, singular-value-weighted scoring, and per-head max aggregation.

\section{Conclusion}
\label{sec:conclusion}
We presented \method, a training-free visual-token pruning method that explicitly conditions on the query without relying on attention weights. 
\method constructs the query--vision interaction matrix induced by standard transformer query--key projections and scores visual tokens using its dominant low-rank subspace, retaining tokens that participate in the main query-relevant interaction modes. 
Across image and video benchmarks and multiple VLMs, \method improves performance under matched token budgets, with the largest gains under aggressive compression. 
These results show that a few dominant SVD components are sufficient to capture much of the task-relevant query--vision structure, establishing low-rank query--vision decomposition as a simple, interpretable, and training-free basis for efficient VLM inference.

\section*{Acknowledgements}

This work was partially supported by Samsung Research America
through the 6G@UT center within the Wireless Networking and
Communications Group (WNCG) at the University of Texas at Austin,
ARO Award W911NF2310062, ONR Award N000142412542,
and NSF CAREER Award 2443857.

\FloatBarrier
\bibliographystyle{abbrvnat}
{\small
\bibliography{references}
}

\newpage

\appendix

\section{Limitations and Future Directions}\label{sec:limitations}
Our evaluation is limited to public VLMs and standard image/video benchmarks, and may not fully capture behavior on closed-source models, proprietary inference stacks, or specialized domains. 
As a training-free pruning method, \method also inherits the biases and failure modes of the underlying VLM. Practical speedups may vary with hardware, batching, sequence length, and attention-kernel implementations.
Future work could extend query--vision decomposition to adaptive token budgets, dynamic selection of the number of SVD components and tighter integration with hardware-aware inference systems. 
It would also be valuable to study whether the same low-rank interaction structure can guide token pruning during training, in longer video settings, or in broader multimodal architectures.

\section{Layer Selection Results}
\label{app:layer_selection}

We provide additional ablations on the choice of pruning layer $\ell$. 
For each model, we compute the QViD scores using intermediate representations from layer $\ell$ and evaluate the resulting compressed model. 
This study tests whether token selection is more effective when computed from the raw input representations or after a few early layers of cross-token processing.

Table~\ref{tab:layer_sweep_llava} reports the layer sweep for LLaVA-1.5-7B at 128 retained tokens. 
Table~\ref{tab:qwen_layer_full} reports the full Qwen2-VL-7B layer sweep across all three retention ratios.

\begin{table*}[t]
\caption{\textbf{Layer Selection Ablation on LLaVA-1.5-7B.}
We vary the scoring layer $\ell$ while retaining 128 visual tokens. 
MME denotes the sum of MME-Perception and MME-Cognition. 
Avg. denotes the mean normalized performance relative to the full-token baseline across all reported metrics. 
\textbf{Bold} denotes the best value in each column.}
\centering
\small
\setlength{\tabcolsep}{5pt}
\renewcommand{\arraystretch}{1.05}
\begin{tabular}{lccccccc}
\toprule
Layer $\ell$ & AI2D & GQA & TextVQA & POPE & MME & SQA & Avg. \\
\midrule
Baseline & 55.2 & 61.9 & 58.2 & 85.9 & 1862.0 & 69.5 & 100.0\% \\
\midrule
1 & 53.7 & 56.7 & 56.7 & 85.1 & 1718.8 & 68.9 & 96.1\% \\
2 & 54.0 & \textbf{57.9} & 56.8 & \textbf{85.6} & 1710.9 & 68.6 & 96.5\% \\
3 & \textbf{54.1} & 57.7 & 57.1 & 85.5 & 1726.3 & 68.7 & \textbf{96.7\%} \\
4 & 53.5 & 57.2 & 57.2 & 85.4 & 1719.3 & \textbf{69.0} & 96.4\% \\
5 & 53.7 & 57.0 & \textbf{57.3} & 85.4 & \textbf{1729.6} & 68.6 & 96.5\% \\
6 & 53.4 & 56.6 & 57.1 & 85.4 & 1678.6 & 68.4 & 95.7\% \\
\bottomrule
\end{tabular}
\label{tab:layer_sweep_llava}
\end{table*}

For LLaVA-1.5-7B, performance is relatively stable across early layers. 
Layer $\ell=3$ achieves the best average performance, but the gap between layers 1--5 is small, suggesting that QViD is not highly sensitive to the exact early layer used for scoring. 
The mild drop at layer 6 suggests that pruning too late may begin to discard visual tokens after task-relevant information has already been mixed across tokens.

\begin{table*}[t]
\caption{\textbf{Full Layer Sweep on Qwen2-VL-7B.}
We vary the QViD scoring layer $\ell$ across three retention ratios. 
Avg. is computed as the mean normalized performance relative to the full-token baseline. 
\textbf{Bold} denotes the best value in each column within a retention block.}
\centering
\small
\setlength{\tabcolsep}{4pt}
\renewcommand{\arraystretch}{1.05}
\begin{tabular}{lccccccccc}
\toprule
Layer $\ell$ & AI2D & GQA & TextVQA & POPE & MME-P & MME-C & SQA & MMB & Avg. \\
\midrule
Baseline & 80.7 & 61.9 & 81.5 & 88.2 & 1684.0 & 635.7 & 85.3 & 84.9 & 100.0\% \\
\midrule
\multicolumn{10}{c}{\textit{Retain 33.3\% Tokens} ($\downarrow 66.7\%$)} \\
\midrule
1 & 74.8 & 57.2 & 78.5 & 84.9 & 1604.6 & 477.9 & 81.4 & 79.1 & 92.1\% \\
2 & 75.7 & 58.0 & \textbf{79.4} & 85.9 & 1618.4 & 470.7 & \textbf{81.7} & 80.8 & 92.9\% \\
3 & \textbf{76.6} & 58.6 & 79.0 & 85.6 & 1610.6 & 493.6 & 81.2 & \textbf{81.0} & 93.4\% \\
4 & 75.9 & \textbf{59.1} & 77.9 & 86.7 & 1596.0 & 502.5 & 81.5 & 80.5 & 93.4\% \\
5 & 74.8 & \textbf{59.1} & 76.4 & \textbf{87.1} & 1608.1 & 459.6 & 81.3 & 78.6 & 92.0\% \\
6 & 74.5 & 57.4 & 78.2 & 85.1 & \textbf{1627.1} & \textbf{543.2} & 80.9 & 80.2 & \textbf{93.6\%} \\
\midrule
\multicolumn{10}{c}{\textit{Retain 22.2\% Tokens} ($\downarrow 77.8\%$)} \\
\midrule
1 & 71.3 & 54.0 & 76.7 & 81.1 & 1504.4 & 426.4 & 78.1 & 73.5 & 87.0\% \\
2 & 72.4 & 55.1 & 76.5 & 83.3 & \textbf{1583.4} & 384.3 & 79.6 & 76.2 & 88.1\% \\
3 & \textbf{73.8} & 56.0 & \textbf{77.1} & 83.7 & 1577.9 & 446.1 & 79.0 & \textbf{78.8} & \textbf{90.1\%} \\
4 & 72.8 & 56.6 & 74.5 & 85.0 & 1519.3 & \textbf{447.9} & \textbf{79.9} & 76.2 & 89.2\% \\
5 & 71.9 & \textbf{57.3} & 71.7 & \textbf{85.6} & 1499.5 & 367.1 & 79.8 & 74.1 & 86.8\% \\
6 & 71.7 & 54.8 & 76.1 & 82.0 & 1556.8 & 426.1 & 78.6 & 75.9 & 88.1\% \\
\midrule
\multicolumn{10}{c}{\textit{Retain 11.1\% Tokens} ($\downarrow 88.9\%$)} \\
\midrule
1 & 66.4 & 47.5 & 68.4 & 73.4 & 1287.9 & 324.6 & 75.1 & 62.1 & 76.9\% \\
2 & 68.9 & 49.1 & 66.8 & 77.0 & 1340.0 & 343.6 & 74.8 & 66.4 & 79.2\% \\
3 & \textbf{69.8} & \textbf{50.9} & \textbf{69.1} & 78.4 & \textbf{1420.1} & \textbf{352.5} & 75.3 & \textbf{70.7} & \textbf{81.7\%} \\
4 & 68.7 & 50.3 & 63.9 & 79.5 & 1343.5 & 322.5 & \textbf{75.6} & 66.1 & 79.0\% \\
5 & 66.5 & \textbf{50.9} & 58.9 & \textbf{80.2} & 1309.8 & 348.9 & 74.0 & 62.3 & 77.6\% \\
6 & 67.9 & 48.7 & 65.6 & 75.7 & 1299.2 & 327.9 & 74.3 & 65.5 & 77.8\% \\
\bottomrule
\end{tabular}

\label{tab:qwen_layer_full}
\end{table*}

For Qwen2-VL-7B, the choice of layer has a more visible effect, especially at lower retention ratios. Layer $\ell=3$ gives the best average performance at 22.2\% and 11.1\% retention, while remaining within 0.2 percentage points of the best average at 33.3\% retention. This suggests that a few early layers help refine the query--vision compatibility scores before pruning, while delaying pruning too much can reduce robustness under more aggressive compression. Overall, these results support using $\ell=3$ as the default pruning layer in our experiments.

\section{Effect of the Number of SVD Components}
\label{app:nsvd_sweep}
\begin{table*}[t]
\centering
\small
\setlength{\tabcolsep}{4.5pt}
\renewcommand{\arraystretch}{1.05}
\begin{tabular}{lcccccc}
\toprule
Method & GQA & SQA & TextVQA & POPE & MME & Avg. \\
\midrule
Baseline & 61.9 & 69.5 & 58.2 & 85.9 & 1862 & 100.0\% \\
\midrule
\multicolumn{7}{c}{\textit{Retain 64 Tokens ($\downarrow$88.9\%)}} \\
\midrule
V$^2$Drop          & 50.5 & \textbf{68.9} & 51.8 & 75.1 & 1470.0 & 87.2\% \\
$n_{\mathrm{svd}}=1$ & 42.1 & 63.8 & 42.6 & 20.7 & 1002.7 & 62.2\% \\
$n_{\mathrm{svd}}=2$ & 47.4 & 67.5 & 48.3 & 54.8 & 1362.3 & 78.7\% \\
$n_{\mathrm{svd}}=4$ & 53.5 & 68.1 & 54.2 & 74.0 & 1599.5 & 89.9\% \\
$n_{\mathrm{svd}}=8$ & \textbf{54.1} & 68.2 & \textbf{55.5} & 76.0 & \textbf{1623.8} & \textbf{91.3\%} \\
$n_{\mathrm{svd}}=16$ & 52.2 & 68.3 & \textbf{55.5} & \textbf{80.8} & 1553.0 & 91.1\% \\
\midrule
\multicolumn{7}{c}{\textit{Retain 128 Tokens ($\downarrow$77.8\%)}} \\
\midrule
V$^2$Drop          & 56.3 & \textbf{68.8} & 53.8 & 80.9 & 1712.0 & 93.7\% \\
$n_{\mathrm{svd}}=1$ & 45.6 & 63.4 & 43.5 & 35.5 & 1077.1 & 67.8\% \\
$n_{\mathrm{svd}}=2$ & 50.3 & 68.3 & 49.6 & 61.7 & 1478.6 & 83.2\% \\
$n_{\mathrm{svd}}=4$ & 56.7 & 68.5 & 56.0 & 80.1 & 1697.0 & 94.2\% \\
$n_{\mathrm{svd}}=8$ & 57.3 & 68.6 & 57.0 & 82.6 & 1725.1 & 95.6\% \\
$n_{\mathrm{svd}}=16$ & \textbf{57.7} & 68.7 & \textbf{57.1} & \textbf{85.5} & \textbf{1726.0} & \textbf{96.5\%} \\
\midrule
\multicolumn{7}{c}{\textit{Retain 192 Tokens ($\downarrow$66.7\%)}} \\
\midrule
V$^2$Drop          & 58.5 & \textbf{69.3} & 55.6 & 85.1 & \textbf{1826.0} & 97.4\% \\
$n_{\mathrm{svd}}=1$ & 48.4 & 63.2 & 43.8 & 50.3 & 1204.7 & 73.5\% \\
$n_{\mathrm{svd}}=2$ & 55.1 & 68.5 & 51.5 & 73.6 & 1612.7 & 89.7\% \\
$n_{\mathrm{svd}}=4$ & 58.2 & 68.9 & 57.1 & 83.1 & 1768.2 & 96.6\% \\
$n_{\mathrm{svd}}=8$ & 59.1 & 68.3 & 57.6 & 84.8 & 1788.2 & 97.5\% \\
$n_{\mathrm{svd}}=16$ & \textbf{59.3} & 69.0 & \textbf{57.8} & \textbf{87.4} & 1793.0 & \textbf{98.5\%} \\
\bottomrule
\end{tabular}
\caption{\textbf{Ablation on the Number of SVD Components for LLaVA-v1.5-7B.}
$n_{\mathrm{svd}}$ denotes the number of singular-vector components used to compute the visual-token importance scores. We vary $n_{\mathrm{svd}}$, the number of singular-vector components used for QViD scoring, across three token budgets. 
Avg. denotes the mean normalized performance relative to the full-token baseline across all reported metrics.}
\label{tab:llava_nsvd_sweep}
\end{table*}
Here, $n_{\mathrm{svd}}$ denotes the number of singular-vector components used to compute the visual-token importance scores. The results show that QViD can capture useful query--vision structure with only a small number of components, but becomes more robust as more dominant interaction modes are included. 
In particular, performance improves sharply from $n_{\mathrm{svd}}=1,2$ to $n_{\mathrm{svd}}=4,8$, showing that single-mode scoring is insufficient under compression. 
The default setting $n_{\mathrm{svd}}=16$ achieves the best average performance at 128 and 192 retained tokens and remains close to the best setting at 64 tokens. 
This justifies using multiple SVD components in the final scoring rule and shows that the dominant query--vision subspace provides a stronger pruning signal than lower-rank alternatives.

\section{Scoring Rule Ablation}
\label{app:scoring_rule_ablation}

We ablate different ways of converting the query--vision interaction matrix into visual-token scores. 
All variants use the same interaction matrix \(M=Q_{\mathrm{txt}}K_{\mathrm{vis}}^\top\), but differ in how token importance is computed from \(M\).

\paragraph{Compared Scoring Rules.} \textbf{Column Norm} directly scores each visual token by the \(\ell_2\)-norm of its corresponding column in \(M\), without using an SVD decomposition.
This measures the raw interaction magnitude between the visual token and the query tokens.
\textbf{Sigma-Weighted} computes an SVD of the global interaction matrix and scores tokens by weighting each right singular-vector component by the corresponding singular value.
\textbf{Per-Head Max} computes SVD-based scores independently for each attention head, and then aggregates scores by taking the maximum across heads.
The default \method scoring uses per-head SVD scores with the fixed aggregation rule used in the main experiments.

\begin{table*}[t]
\caption{\textbf{Scoring-Rule Ablation on LLaVA-1.5-7B.}
We compare different scoring rules derived from the query--vision interaction matrix at matched token budgets. 
Avg. denotes the mean normalized performance relative to the full-token baseline across all reported metrics.}
\centering
\small
\setlength{\tabcolsep}{4.5pt}
\renewcommand{\arraystretch}{1.05}
\begin{tabular}{lcccccc}
\toprule
Method & GQA & SQA & TextVQA & POPE & MME & Avg. \\
\midrule
Baseline & 61.9 & 69.5 & 58.2 & 85.9 & 1862.0 & 100.0\% \\
\midrule
\multicolumn{7}{c}{\textit{Retain 64 Tokens ($\downarrow$88.9\%)}} \\
\midrule
Column norm     & 42.2 & 64.0 & 42.6 & 54.7 & 1009.0 & 70.3\% \\
Sigma-weighted  & 42.2 & 64.0 & 42.6 & 54.7 & 1009.0 & 70.3\% \\
Per-head max    & \textbf{53.2} & 68.1 & \textbf{55.7} & \textbf{82.1} & \textbf{1581.0} & \textbf{92.0\%} \\
\method       & 52.2 & \textbf{68.3} & 55.5 & 80.8 & 1553.0 & 91.1\% \\
\midrule
\multicolumn{7}{c}{\textit{Retain 128 Tokens ($\downarrow$77.8\%)}} \\
\midrule
Column norm     & 45.3 & 63.7 & 43.4 & 59.3 & 1069.0 & 73.2\% \\
Sigma-weighted  & 45.3 & 63.7 & 43.4 & 59.3 & 1070.0 & 73.2\% \\
Per-head max    & 57.6 & 68.4 & \textbf{57.1} & \textbf{86.3} & \textbf{1734.0} & \textbf{96.6\%} \\
\method       & \textbf{57.7} & \textbf{68.7} & \textbf{57.1} & 85.5 & 1726.0 & 96.5\% \\
\midrule
\multicolumn{7}{c}{\textit{Retain 192 Tokens ($\downarrow$66.7\%)}} \\
\midrule
Column norm     & 47.8 & 63.0 & 43.6 & 65.0 & 1190.0 & 76.5\% \\
Sigma-weighted  & 47.8 & 63.0 & 43.6 & 65.0 & 1190.0 & 76.5\% \\
Per-head max    & \textbf{59.4} & 67.9 & \textbf{58.0} & 87.2 & 1780.0 & 98.1\% \\
\method       & 59.3 & \textbf{69.0} & 57.8 & \textbf{87.4} & \textbf{1793.0} & \textbf{98.5\%} \\
\bottomrule
\end{tabular}

\label{tab:scoring_rule_ablation}
\end{table*}

Table~\ref{tab:scoring_rule_ablation} shows that raw interaction strength alone is not sufficient for reliable token selection. 
Column-norm and sigma-weighted scoring perform identically because they induce the same token ordering. 
Indeed, if \(M=U\Sigma V^\top\), then
\[
    M^\top M = V\Sigma^2 V^\top.
\]
Therefore, for visual token \(j\),
\[
    \|M_{:,j}\|_2^2 
    =
    (M^\top M)_{jj}
    =
    \sum_k \sigma_k^2 V_{jk}^2.
\]
Thus, sigma-weighted scoring is exactly the squared column-norm score. 
Since squaring is monotone for nonnegative scores, both methods select the same top-\(k\) visual tokens and give identical evaluation results. 
Their poor performance shows that selecting tokens by raw interaction energy is much weaker than selecting tokens by their participation in the dominant query--vision subspace.

In contrast, per-head SVD-based scoring performs substantially better, indicating that the main benefit comes from preserving head-specific low-rank query--vision structure rather than simply selecting tokens with large interaction magnitude.

The per-head max variant is competitive with the default \method aggregation and performs slightly better on some metrics, especially at smaller token budgets. This suggests that different attention heads may specialize in different query-relevant visual cues. In this work, we keep the default aggregation fixed for consistency across models and budgets. 
Adaptive or query-dependent head aggregation is an interesting direction for future work.

\section{Efficient \method Scoring}
\label{app:efficient_qr_svd}

\method scores visual tokens using the dominant right singular directions of the query--vision interaction matrix. 
For each attention head \(h\), we form
\begin{equation*}
    M^{(h)}
    =
    Q_{\mathrm{txt}}^{(h)}
    \left(K_{\mathrm{vis}}^{(h)}\right)^\top
    \in \mathbb{R}^{N_q \times N_v},
\end{equation*}
where \(N_q\) is the number of query tokens and \(N_v\) is the number of visual tokens. 
A direct implementation could compute the SVD of \(M^{(h)}\) and use its top right singular vectors for token scoring. 
However, since \(N_q \ll N_v\), we can compute the same right-singular subspace more efficiently using a QR decomposition.

\paragraph{QR-Based Reduction.}
Consider the transpose
\[
    \left(M^{(h)}\right)^\top \in \mathbb{R}^{N_v \times N_q},
\]
which is tall and thin. 
We first compute its thin QR decomposition:
\begin{equation*}
    \left(M^{(h)}\right)^\top
    =
    \widehat{Q}^{(h)} R^{(h)},
    \qquad
    \widehat{Q}^{(h)} \in \mathbb{R}^{N_v \times N_q},
    \quad
    R^{(h)} \in \mathbb{R}^{N_q \times N_q},
\end{equation*}
where the columns of \(\widehat{Q}^{(h)}\) are orthonormal and \(R^{(h)}\) is upper triangular. 
We then compute the SVD of the small matrix \(\left(R^{(h)}\right)^\top\):
\begin{equation*}
    \left(R^{(h)}\right)^\top
    =
    \widetilde{U}^{(h)}
    \Sigma^{(h)}
    \left(\widetilde{V}^{(h)}\right)^\top
\end{equation*}

Substituting the QR factorization into \(M^{(h)}\), we obtain
\begin{align*}
    M^{(h)}
    &=
    \left(R^{(h)}\right)^\top
    \left(\widehat{Q}^{(h)}\right)^\top \\
    &=
    \widetilde{U}^{(h)}
    \Sigma^{(h)}
    \left(\widetilde{V}^{(h)}\right)^\top
    \left(\widehat{Q}^{(h)}\right)^\top \\
    &=
    \widetilde{U}^{(h)}
    \Sigma^{(h)}
    \left(
    \widehat{Q}^{(h)}
    \widetilde{V}^{(h)}
    \right)^\top
\end{align*}
Therefore, the right singular vectors of \(M^{(h)}\) are
\begin{equation*}
    V^{(h)}
    =
    \widehat{Q}^{(h)}
    \widetilde{V}^{(h)}
    \in \mathbb{R}^{N_v \times N_q}
\end{equation*}
Thus, the QR-based computation is mathematically equivalent to computing the right singular vectors of \(M^{(h)}\) directly, but avoids applying SVD to the full rectangular \(N_q \times N_v\) matrix.

\paragraph{Token Scoring.}
Let \(V^{(h)}_{r_{\mathrm{eff}}}\in\mathbb{R}^{N_v\times r_{\mathrm{eff}}}\) denote the first \(r_{\mathrm{eff}}\) columns of \(V^{(h)}\). 
In our experiments, we set \(r=n_{\mathrm{svd}}\) and use
\begin{equation*}
    r_{\mathrm{eff}}=\min(n_{\mathrm{svd}},N_q,N_v),
\end{equation*}
so the score is well defined even for short queries. 
The per-head visual-token score is the row-leverage score
\begin{equation*}
    s_j^{(h)}
    =
    \sum_{k=1}^{r_{\mathrm{eff}}}
    \left(V_{jk}^{(h)}\right)^2
\end{equation*}
We average these scores across attention heads:
\begin{equation*}
    s_j
    =
    \frac{1}{H}
    \sum_{h=1}^{H}
    s_j^{(h)}
\end{equation*}
The top-\(K\) visual tokens by \(s_j\) are retained and passed to the remaining LLM layers.

\paragraph{Complexity.}
For each head, the QR decomposition of \(\left(M^{(h)}\right)^\top\in\mathbb{R}^{N_v\times N_q}\) costs \(O(N_vN_q^2)\), and the SVD of the small \(N_q\times N_q\) matrix costs \(O(N_q^3)\). 
Across \(H\) heads, the scoring complexity is
\begin{equation*}
    O(HN_vN_q^2 + HN_q^3)
\end{equation*}
Since \(N_q\) is typically much smaller than \(N_v\), this cost is small compared with the cost of forwarding all visual tokens through the LLM. 
In our LLaVA-1.5-7B experiments, the full scoring step, including construction of \(M=Q_{\mathrm{txt}}K_{\mathrm{vis}}^\top\), costs less than \(50\) MFLOPs.

\paragraph{Implementation.}
The QR-based scoring step can be implemented compactly as follows:
\begin{verbatim}
# M: (H, N_q, N_v), n_svd = number of SVD components
r_eff = min(n_svd, M.shape[-2], M.shape[-1])

Q_qr, R_qr = torch.linalg.qr(M.mT)
_, _, Vh = torch.linalg.svd(R_qr.mT, full_matrices=False)

# Right singular vectors of M
W = (Q_qr @ Vh.mT)[:, :, :r_eff]  # (H, N_v, r_eff)

# Visual-token leverage scores, averaged over heads
scores = (W * W).sum(dim=-1).mean(dim=0)  # (N_v,)
\end{verbatim}

\section{Detailed Efficiency Results}
\label{app:efficiency}

In this section, we provide detailed latency, throughput, and FLOPs measurements for LLaVA-1.5-7B. 
The main paper reports the high-level compute comparison. 
Here, we include the full efficiency breakdown used for the complexity analysis.

\paragraph{Scoring Overhead.} \method adds a lightweight query--vision scoring pass before pruning: with the QR-based implementation described in Appendix~\ref{app:efficient_qr_svd}, the scoring cost is
\(O(HN_vN_q^2 + HN_q^3)\) for \(H\) attention heads, \(N_q\) query tokens, and \(N_v\) visual tokens. Since \(N_q \ll N_v\) in typical VLM inputs, this overhead is small relative to forwarding all visual tokens through the LLM and is amortized by pruning; detailed derivations and efficiency measurements are provided in Appendices~\ref{app:efficient_qr_svd} as well as Section~\ref{sec:results}. 

\paragraph{Compute Accounting.}
For the full-token baseline, the model processes all \(576\) visual tokens. 
For V$^2$Drop, the reported TFLOPs include its per-layer token schedule, where early layers retain the full visual sequence before pruning. 
For \method, the reported TFLOPs include the early query-aware scoring pass and the subsequent LLM computation using the retained visual tokens. 
The QR/SVD scoring computation itself is small, costing less than \(50\) MFLOPs including construction of the interaction matrix.

From our FLOP measurements on LLaVA-1.5-7B, removing one visual token from the remaining LLM computation saves roughly \(0.01\) TFLOPs. 
Therefore, retaining \(64\), \(128\), and \(192\) tokens instead of all \(576\) visual tokens saves approximately \(7.17\), \(6.27\), and \(5.38\) TFLOPs, respectively, in the remaining LLM computation. 
This shows that the \method scoring overhead is much smaller than the compute saved by pruning.

\begin{table}[t]
\centering
\tiny
\setlength{\tabcolsep}{5pt}
\renewcommand{\arraystretch}{1.05}
\caption{\textbf{Detailed Efficiency Results on LLaVA-1.5-7B.}
We report measured latency, throughput, speedup, and total TFLOPs for different pruning methods and retained-token budgets. 
TFLOPs include pruning overhead.}
\label{tab:detailed_efficiency_llava_appendix}
\resizebox{0.85\linewidth}{!}{
\begin{tabular}{lccccc}
\toprule
Method & Tokens & Lat. (ms) & Tput. (img/s) & Speedup & TFLOPs \\
\midrule
Baseline & 576 & 100.1 & 8.60 & 1.00$\times$ & 8.86 \\
\midrule
\method & 64  & 80.1 & 10.40 & 1.21$\times$ & 2.53 \\
\method & 128 & 81.3 & 10.27 & 1.19$\times$ & 3.42 \\
\method & 192 & 81.5 & 10.26 & 1.19$\times$ & 4.32 \\
\midrule
V$^2$Drop & 64  & 77.7 & 10.66 & 1.24$\times$ & 1.97 \\
V$^2$Drop & 128 & 78.6 & 10.55 & 1.23$\times$ & 2.85 \\
V$^2$Drop & 192 & 78.2 & 10.58 & 1.23$\times$ & 3.73 \\
\bottomrule
\end{tabular}
}
\end{table}

\paragraph{Latency and Throughput.}
\method has slightly higher latency than V$^2$Drop at the same retained-token budget because it computes query-aware scores before pruning. 
However, it still improves throughput over the full-token baseline. 
The full baseline runs at \(8.60\) images/s, while \method runs at \(10.40\), \(10.27\), and \(10.26\) images/s for \(64\), \(128\), and \(192\) retained tokens, respectively. 
Thus, \method introduces modest additional overhead relative to V$^2$Drop, but remains substantially more efficient than the full-token baseline.
\begin{figure}
    \centering
    \includegraphics[width=\linewidth]{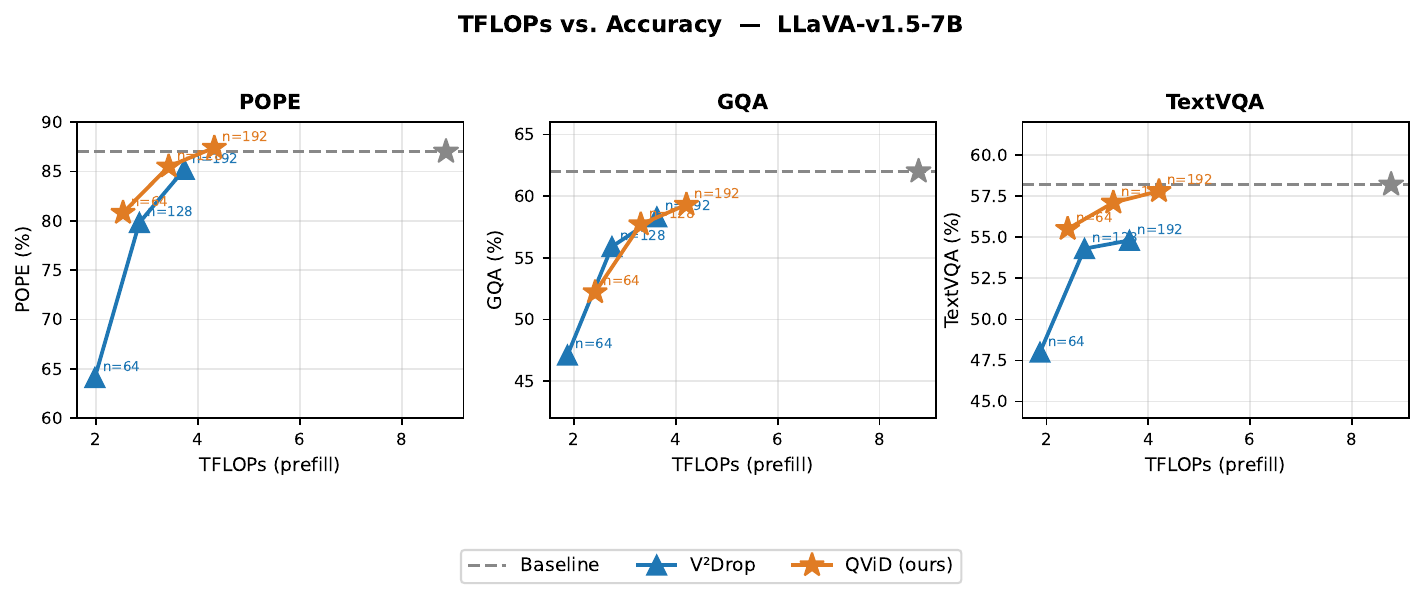}
    \caption{\textbf{Accuracy--Compute Tradeoff on LLaVA-1.5-7B.}
    We compare V$^2$Drop, and \method at 64, 128, and 192 retained visual tokens on POPE, GQA, and TextVQA. 
    The gray dashed line and star indicate the full-token baseline with 576 visual tokens. 
    \method achieves a stronger accuracy--compute tradeoff, especially under aggressive pruning, by retaining more query-relevant visual tokens at comparable TFLOPs.}
    \label{fig:tflops_accuracy_llava}
\end{figure}

\paragraph{Implementation details.}
All experiments are conducted using Python 3.11, PyTorch 2.5.1, and CUDA 12.1 on a single NVIDIA A100-80GB GPU. For LLaVA-1.5-7B, Qwen2-VL-7B, Video-LLaVA-7B, and LLaVA-OV-7B, we follow the standard evaluation settings of the corresponding model and benchmark. Unless otherwise stated, QViD prunes at LLM layer \(\ell=3\) using \(n_{\mathrm{svd}}=16\) dominant SVD components, and omits Rotary Positional Embedding (RoPE) when constructing the query--vision interaction matrix. Reported TFLOPs include both the pruning overhead and the subsequent LLM computation on the retained visual tokens.

\section{Additional Multimodal Benchmark Results}
\label{app:additional_multimodal_benchmarks}

We further evaluate \method on additional multimodal benchmarks beyond the main evaluation suite. 
For LLaVA-v1.5-7B, we report results on AI2D, ChartQA, MMMU, OCRBench, and SEEDBench. 
For Qwen2-VL-7B, we additionally evaluate on MMMU-Pro Standard, a more challenging multimodal reasoning benchmark with 10 answer options.

\begin{table*}[t]
\centering
\small
\setlength{\tabcolsep}{5pt}
\renewcommand{\arraystretch}{1.05}
\begin{tabular}{lccccccc}
\toprule
Method & $n$ & AI2D & ChartQA & MMMU & OCRBench & SEEDBench & Avg. \\
\midrule
Baseline & 576 & 55.2 & 18.2 & 36.1 & 31.5 & 66.2 & 100.0\% \\
\midrule
\multicolumn{8}{c}{\textit{Retain 64 Tokens} ($\downarrow 88.9\%$)} \\
\midrule
V$^2$Drop & 64 & \textbf{52.3} & 13.2 & 33.3 & 14.4 & 48.0 & 75.5\% \\
\method   & 64 & 52.2 & \textbf{16.6} & \textbf{35.0} & \textbf{26.5} & \textbf{54.6} & \textbf{89.9\%} \\
\midrule
\multicolumn{8}{c}{\textit{Retain 128 Tokens} ($\downarrow 77.8\%$)} \\
\midrule
V$^2$Drop & 128 & 53.5 & 15.0 & \textbf{35.1} & 26.2 & 59.4 & 89.9\% \\
\method   & 128 & \textbf{54.1} & \textbf{17.0} & \textbf{35.1} & \textbf{30.5} & \textbf{61.0} & \textbf{95.5\%} \\
\midrule
\multicolumn{8}{c}{\textit{Retain 192 Tokens} ($\downarrow 66.7\%$)} \\
\midrule
V$^2$Drop & 192 & \textbf{54.7} & 15.8 & 35.7 & 26.8 & \textbf{63.8} & 93.3\% \\
\method   & 192 & 54.3 & \textbf{17.4} & \textbf{37.1} & \textbf{30.9} & 63.5 & \textbf{98.2\%} \\
\bottomrule
\end{tabular}
\caption{\textbf{Additional Multimodal Benchmark Results on LLaVA-v1.5-7B.}
We compare \method with V$^2$Drop at matched token budgets on AI2D, ChartQA, MMMU, OCRBench, and SEEDBench.
Avg. denotes the mean normalized performance relative to the full-token baseline.}
\label{tab:llava_additional_benchmarks}
\end{table*}

\begin{table}[t]
\caption{\textbf{MMMU-Pro Standard Results on Qwen2-VL-7B.}
We compare \method with V$^2$Drop under matched retention ratios on the 1,730-sample standard-format split.}
\label{tab:qwen_mmmupro}
\centering
\small
\setlength{\tabcolsep}{8pt}
\renewcommand{\arraystretch}{1.0}
\begin{tabular}{@{}lcc@{}}
\toprule
Method & Retention & MMMU-Pro \\
\midrule
Baseline & 100\% & 33.8 \\
\midrule
V$^2$Drop & 33.3\% & \textbf{32.3} \\
\method   & 33.3\% & 31.7 \\
\midrule
V$^2$Drop & 22.2\% & \textbf{30.2} \\
\method   & 22.2\% & 29.9 \\
\midrule
V$^2$Drop & 11.1\% & 23.6 \\
\method   & 11.1\% & \textbf{28.8} \\
\bottomrule
\end{tabular}
\end{table}

Table~\ref{tab:llava_additional_benchmarks} shows that \method consistently improves over V$^2$Drop in average normalized performance across additional LLaVA-v1.5-7B benchmarks. 
The gains are especially large on OCRBench and ChartQA, suggesting that query-aware token selection preserves visually grounded information beyond the main evaluation suite. 

Table~\ref{tab:qwen_mmmupro} reports results on MMMU-Pro Standard for Qwen2-VL-7B. 
In the standard-format split, the model receives the question, answer choices, and associated image as structured multimodal input. 
\method remains close to V$^2$Drop at 33.3\% and 22.2\% retention, while substantially improving performance at 11.1\% retention, from 23.6\% to 28.8\%. 
This indicates that query-aware token selection is especially beneficial under aggressive compression on challenging multimodal reasoning tasks.

\section{Additional Video Understanding Results}
\label{app:video_llava_results}

Table~\ref{tab:video_llava_results_appendix} reports additional video understanding results on Video-LLaVA-7B. 
The trends are consistent with the LLaVA-OV-7B results in the main paper: \method outperforms V$^2$Drop at all retention ratios, with the largest gain under the most aggressive compression setting.

\begin{table}[t]
\caption{\textbf{Video Understanding Results on Video-LLaVA-7B.}
We compare \method with V$^2$Drop on MVBench and VideoMME under token compression.
Avg. denotes the mean normalized performance over MVBench and VideoMME Overall relative to the full-token baseline.}
\label{tab:video_llava_results_appendix}
\centering
\footnotesize
\setlength{\tabcolsep}{4.2pt}
\renewcommand{\arraystretch}{0.96}
\begin{tabular}{@{}lcccccc@{}}
\toprule
Method & MVBench & \multicolumn{4}{c}{VideoMME} & Avg. \\
\cmidrule(lr){3-6}
& & Overall & Short & Med. & Long & \\
\midrule
\multicolumn{7}{l}{\textit{All Tokens (100\%)}} \\
Baseline & 42.9 & 36.2 & 41.2 & 35.0 & 32.4 & 100.0\% \\
\midrule

\multicolumn{7}{c}{\textit{Retention = 25\%}} \\
\method & \textbf{41.5} & \textbf{37.7} & \textbf{43.2} & \textbf{36.1} & \textbf{33.8} & \textbf{100.4\%} \\
V$^2$Drop & 41.2 & 36.3 & 40.1 & 35.1 & 33.7 & 98.2\% \\
\midrule

\multicolumn{7}{c}{\textit{Retention = 15\%}} \\
\method & \textbf{39.8} & \textbf{35.6} & \textbf{38.7} & \textbf{34.4} & \textbf{33.8} & \textbf{95.6\%} \\
V$^2$Drop & 38.4 & 32.8 & 35.3 & 32.3 & 30.8 & 90.1\% \\
\midrule

\multicolumn{7}{c}{\textit{Retention = 10\%}} \\
\method & \textbf{39.5} & \textbf{34.7} & \textbf{36.9} & \textbf{34.1} & \textbf{33.2} & \textbf{94.0\%} \\
V$^2$Drop & 33.6 & 29.9 & 29.4 & 29.8 & 30.4 & 80.5\% \\
\bottomrule
\end{tabular}
\end{table}

\section{Hallucination and Captioning Results}
\label{app:hallusion_coco}

We further evaluate \method on HallusionBench and COCO Captioning to test whether token compression preserves both hallucination robustness and open-ended generation quality. 
HallusionBench is a hallucination-sensitive benchmark and is reported using all-pair accuracy (aAcc), figure accuracy (fAcc), and question accuracy (qAcc). 
COCO Captioning is evaluated using CIDEr, a caption-quality metric based on consensus between generated captions and human reference captions. 
CIDEr is not an accuracy metric; in our lmms-eval setup it is reported on a normalized scale, where values around 1.0 roughly correspond to a traditional CIDEr score around 100. 
Higher is better for all metrics.

\begin{table*}[t]
\caption{\textbf{HallusionBench and COCO Captioning Results on LLaVA-v1.5-7B.}
We compare \method with V$^2$Drop at matched token budgets.
HallusionBench is reported using aAcc, fAcc, and qAcc, while COCO Captioning is reported using CIDEr.
\method improves both hallucination-related metrics and caption quality under visual-token compression.}
\centering
\small
\setlength{\tabcolsep}{6pt}
\renewcommand{\arraystretch}{1.05}
\begin{tabular}{lccccc}
\toprule
Method & $n$ & \multicolumn{3}{c}{HallusionBench} & COCO Captioning \\
\cmidrule(lr){3-5} \cmidrule(lr){6-6}
& & aAcc & fAcc & qAcc & CIDEr \\
\midrule
Baseline & 576 & 47.9 & 20.5 & 12.3 & 1.087 \\
\midrule
\multicolumn{6}{c}{\textit{Retain 64 Tokens} ($\downarrow 88.9\%$)} \\
\midrule
V$^2$Drop & 64 & 46.8 & 15.0 & 9.5 & 0.591 \\
\method   & 64 & \textbf{47.8} & \textbf{19.7} & \textbf{12.7} & \textbf{0.782} \\
\midrule
\multicolumn{6}{c}{\textit{Retain 128 Tokens} ($\downarrow 77.8\%$)} \\
\midrule
V$^2$Drop & 128 & 47.1 & 17.6 & 11.9 & 0.939 \\
\method   & 128 & \textbf{48.4} & \textbf{20.8} & \textbf{12.7} & \textbf{0.967} \\
\midrule
\multicolumn{6}{c}{\textit{Retain 192 Tokens} ($\downarrow 66.7\%$)} \\
\midrule
V$^2$Drop & 192 & 47.3 & 18.2 & 11.6 & 0.995 \\
\method   & 192 & \textbf{48.6} & \textbf{20.8} & \textbf{12.7} & \textbf{1.028} \\
\bottomrule
\end{tabular}

\label{tab:llava_hallusion_coco}
\end{table*}

\begin{table*}[t]
\caption{\textbf{HallusionBench and COCO Captioning Results on Qwen2-VL-7B.}
We compare \method with V$^2$Drop at matched retention ratios where available.
HallusionBench is reported using aAcc, fAcc, and qAcc, while COCO Captioning is reported using CIDEr.
\method preserves stronger hallucination robustness and substantially better captioning quality under compression.}
\centering
\small
\setlength{\tabcolsep}{6pt}
\renewcommand{\arraystretch}{1.05}
\begin{tabular}{lccccc}
\toprule
Method & Retention & \multicolumn{3}{c}{HallusionBench} & COCO Captioning \\
\cmidrule(lr){3-5} \cmidrule(lr){6-6}
& & aAcc & fAcc & qAcc & CIDEr \\
\midrule
Baseline & 100\% & 67.5 & 38.7 & 42.2 & 0.977 \\
\midrule
\multicolumn{6}{c}{\textit{Retain 33.3\% Tokens} ($\downarrow 66.7\%$)} \\
\midrule
V$^2$Drop & 33.3\% & 61.8 & 33.8 & 33.6 & 0.563 \\
\method   & 33.3\% & \textbf{62.3} & \textbf{35.3} & \textbf{35.2} & \textbf{0.922} \\
\midrule
\multicolumn{6}{c}{\textit{Retain 22.2\% Tokens} ($\downarrow 77.8\%$)} \\
\midrule
V$^2$Drop & 22.2\% & 57.0 & 27.7 & 25.9 & 0.554 \\
\method   & 22.2\% & \textbf{60.3} & \textbf{31.2} & \textbf{31.2} & \textbf{0.880} \\
\midrule
\multicolumn{6}{c}{\textit{Retain 11.1\% Tokens} ($\downarrow 88.9\%$)} \\
\midrule
\method   & 11.1\% & 58.1 & 28.0 & 27.3 & 0.752 \\
\bottomrule
\end{tabular}

\label{tab:qwen_hallusion_coco}
\end{table*}

Tables~\ref{tab:llava_hallusion_coco} and~\ref{tab:qwen_hallusion_coco} show that \method remains effective on hallucination-sensitive and generative benchmarks. 
On LLaVA-v1.5-7B, \method improves over V$^2$Drop across all HallusionBench metrics and CIDEr at every token budget. 
On Qwen2-VL-7B, \method also improves HallusionBench performance and gives substantially higher CIDEr at matched retention ratios. 
These results suggest that query-aware token selection preserves visual information needed for both grounded recognition and open-ended caption generation.

\section{Additional Qualitative Analysis}
\label{app:additional_qualitative}

\subsection{Low-Rank Structure of Query--Vision Interactions}
\label{app:analysis_spectrum}

        \begin{figure*}[t]
            \centering
            \includegraphics[width=0.75\textwidth]{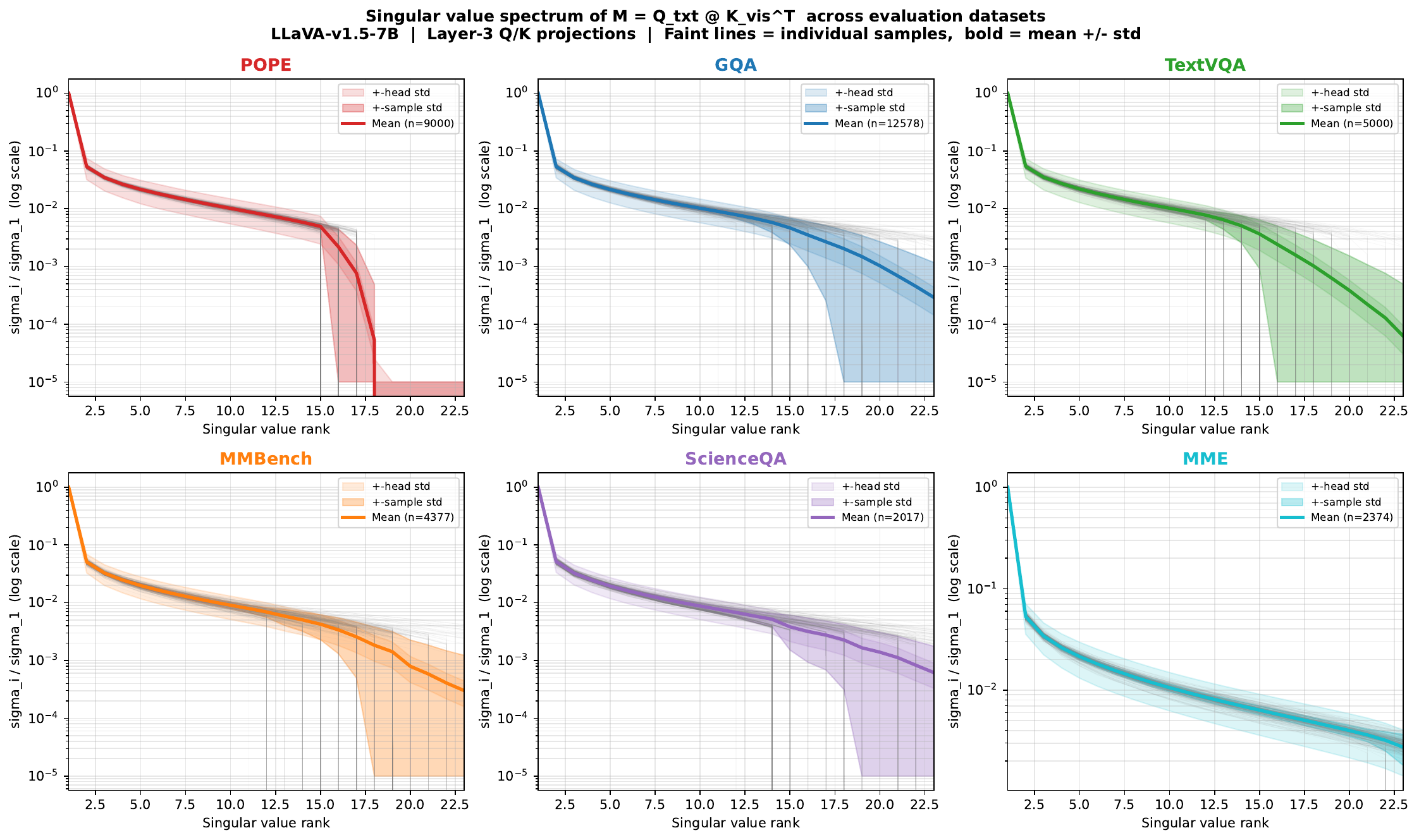}
            \caption{\textbf{Low-Rank Structure of Query--Vision Interactions.}
            We plot the normalized singular-value spectrum \(\sigma_k/\sigma_1\) of \(M=Q_{\mathrm{txt}}K_{\mathrm{vis}}^\top\) at the pruning layer of LLaVA-1.5-7B. 
            Across datasets, the spectra decay rapidly, showing that query--vision interactions are concentrated in a few dominant directions. 
            Faint curves show individual samples, solid curves show the mean, and shaded bands indicate variation across heads.}
            \label{fig:svd_spectrum_datasets}
        \end{figure*}

A key motivation for \method is that the query--vision interaction matrix \(M\) has a concentrated singular spectrum. 
To verify this, we compute the singular values of \(M=Q_{\mathrm{txt}}K_{\mathrm{vis}}^\top\) at the pruning layer across multiple evaluation datasets. 
Figure~\ref{fig:svd_spectrum_datasets} plots the normalized spectrum \(\sigma_k/\sigma_1\) on a logarithmic scale.

Across all datasets, the spectrum decays rapidly after the first few components, showing that most of the query--vision interaction energy is concentrated in a small number of dominant directions. 
This supports our design choice of using only the top SVD components for token scoring. 
The trend is consistent across POPE, GQA, TextVQA, MMBench, ScienceQA, and MME, suggesting that the low-rank structure is not specific to a single benchmark or task type. 
Although the exact decay rate varies across datasets, the dominant components consistently capture the strongest query-conditioned interaction modes.

\subsection{Qualitative Analysis of SVD Components}
\label{app:qualitative_svd_components}
\begin{figure}[t]
    \centering
    \includegraphics[width=0.95\linewidth]{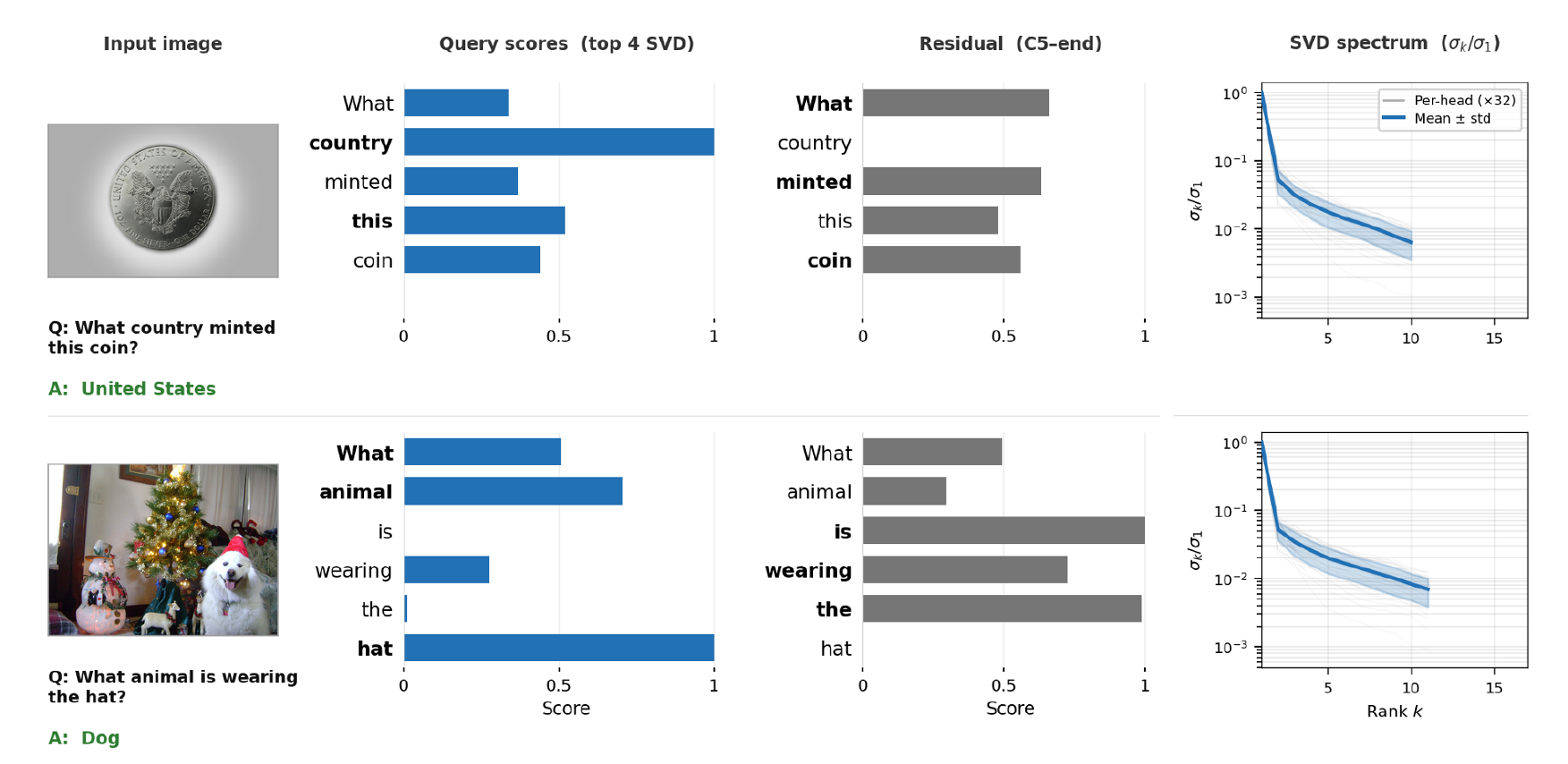}
    \caption{\textbf{Dominant SVD Components Capture Task-Relevant Structure.}
    Top SVD components emphasize task-critical query tokens, while residual components are less aligned with the question. The spectra show that query--vision interactions are concentrated in a few dominant directions.}
    \label{fig:query_score_components}
\end{figure}

For the query-token visualizations, we use the analogous left-singular-vector score. 
If \(M=U\Sigma V^\top\) and \(U_r\) contains the top \(r\) left singular vectors, the query-token score is
$
    q_i = \sum_{k=1}^{r} U_{ik}^{2}.
$
Figure~\ref{fig:query_score_components} compares scores from the dominant SVD components with scores from the residual components obtained after removing the top directions. 
The dominant components assign high scores to task-critical words, such as \emph{country}, \emph{coin}, \emph{animal}, and \emph{hat}, which are directly tied to the answer. 
In contrast, the residual components place more weight on less informative or function words, indicating weaker alignment with the task. 
The singular-value spectra in the same figure also show rapid decay, suggesting that the main query--vision interaction structure is concentrated in a few dominant directions. This supports our choice of using only the dominant SVD components for visual-token scoring.

\section{Confidence Intervals}
\label{app:confidence_intervals}

To assess the statistical reliability of the reported results, we provide 95\% confidence intervals for Qwen2-VL-7B in Table~\ref{tab:ci_qwen2vl_7b}, computed over three independent random seeds. Across different token retention ratios, \method consistently outperforms FastV and V$^2$Drop on GQA and TextVQA, with particularly large improvements at lower retention ratios. At 5.56\% retention, \method also achieves substantial gains across all six benchmarks compared to both baselines.
\begin{table*}[t]
\centering
\scriptsize
\setlength{\tabcolsep}{3.2pt}
\resizebox{\textwidth}{!}{%
\begin{tabular}{clcccccc}
\toprule
Retention & Method & POPE & GQA & TextVQA &
MMStar & AI2D & MME \\
\midrule

-- & Baseline
& $87.3{\pm}0.7$ & $61.9{\pm}0.8$ & $81.2{\pm}1.0$
& $57.6{\pm}2.5$ & $80.7{\pm}1.4$
& $2320{\pm}113$ \\
\midrule

\multirow{3}{*}{5.56\%}
& \textbf{QViD}
& $59.9{\pm}0.9$ & $45.2{\pm}0.9$ & $64.0{\pm}1.3$
& $34.8{\pm}2.4$ & $66.6{\pm}1.7$
& $1520{\pm}153$ \\
& FastV
& $46.3{\pm}1.0$ & $37.7{\pm}0.8$ & $42.5{\pm}1.3$
& $25.7{\pm}2.2$ & $63.7{\pm}1.7$
& $1076{\pm}159$ \\
& V$^2$Drop
& $57.6{\pm}1.0$ & $39.7{\pm}0.9$ & $44.9{\pm}1.3$
& $27.8{\pm}2.3$ & $64.7{\pm}1.7$
& $1090{\pm}159$ \\
\midrule

\multirow{3}{*}{11.1\%}
& \textbf{QViD}
& $73.5{\pm}0.8$ & $50.7{\pm}0.9$ & $72.2{\pm}1.2$
& $39.0{\pm}2.5$ & $69.5{\pm}1.6$
& $1771{\pm}148$ \\
& FastV
& $58.0{\pm}1.0$ & $43.2{\pm}0.9$ & $55.3{\pm}1.3$
& $33.5{\pm}2.4$ & $64.8{\pm}1.7$
& $1479{\pm}149$ \\
& V$^2$Drop
& $57.7{\pm}1.0$ & $39.7{\pm}0.9$ & $45.0{\pm}1.3$
& $28.0{\pm}2.3$ & $65.0{\pm}1.7$
& $1158{\pm}159$ \\
\midrule

\multirow{3}{*}{22.2\%}
& \textbf{QViD}
& $81.3{\pm}0.8$ & $56.2{\pm}0.9$ & $77.3{\pm}1.1$
& $44.6{\pm}2.5$ & $73.8{\pm}1.6$
& $2056{\pm}134$ \\
& FastV
& $74.9{\pm}0.8$ & $51.1{\pm}0.9$ & $71.0{\pm}1.2$
& $41.6{\pm}2.5$ & $69.0{\pm}1.6$
& $1856{\pm}141$ \\
& V$^2$Drop
& $77.8{\pm}0.8$ & $49.6{\pm}0.9$ & $52.1{\pm}1.3$
& $43.4{\pm}2.5$ & $71.9{\pm}1.6$
& $2046{\pm}138$ \\
\bottomrule
\end{tabular}%
}
\caption{95\% confidence intervals on Qwen2-VL-7B over three independent random seeds.}
\label{tab:ci_qwen2vl_7b}
\end{table*}

\subsection{Additional Token-Score Visualizations}
\label{app:additional_visualizations}

\begin{figure*}[p]
    \centering
    \includegraphics[height=0.92\textheight]{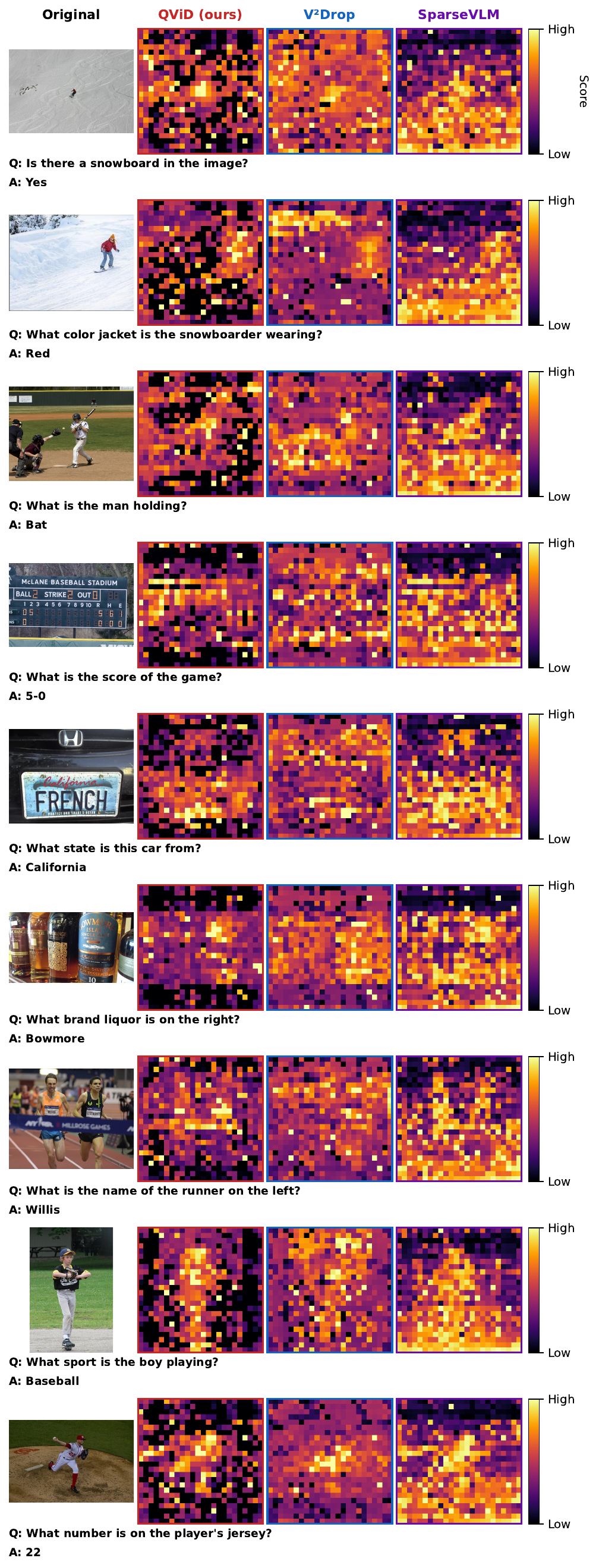}
    \caption{\textbf{Additional Qualitative Token-Score Visualizations.}
    We compare token-score maps from \method, V$^2$Drop, and SparseVLM across multiple image-query pairs. 
    Brighter regions indicate higher token scores. 
    \method tends to assign high scores to query-relevant visual regions, while the baseline methods often produce more diffuse score maps.}
    \label{fig:appendix_score_maps}
\end{figure*}

Figure~\ref{fig:appendix_score_maps} shows additional token-score maps comparing \method with V$^2$Drop and SparseVLM. 
Across diverse image-query pairs, \method assigns high scores to localized regions that are relevant to the question, such as the snowboarder, bat, scoreboard, license plate, and player jersey. 
In contrast, the baseline score maps are often more diffuse or place high scores on less query-relevant regions. 
These examples provide qualitative evidence that query--vision SVD scoring better aligns token importance with the visual evidence needed to answer the input query.

\subsection{Query-Token Scores from Dominant and Residual Components}
\label{app:query_score_components}

\begin{figure*}[p]
    \centering
    \includegraphics[height=0.92\textheight]{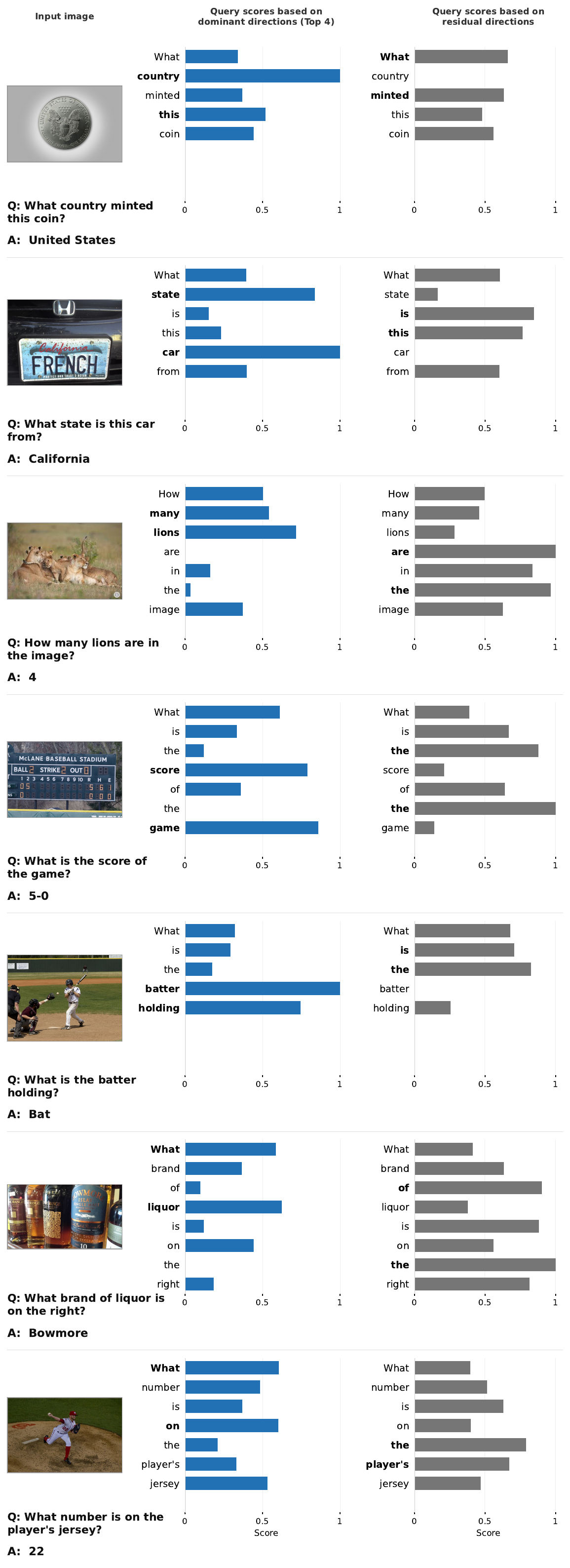}
    \caption{\textbf{Additional Query-Token Score Visualizations.}
    For each example, we compare query-token scores obtained from the dominant SVD components with scores from the residual components. 
    The dominant components place higher weight on semantically informative query tokens, while residual components are less aligned with the task-relevant words.}
    \label{fig:appendix_query_score_components}
\end{figure*}

Figure~\ref{fig:appendix_query_score_components} provides additional examples comparing query-token scores from the dominant SVD components and the residual components. 
For each image-query pair, we compute scores using the top SVD components and compare them with scores from the residual directions. 
The dominant components tend to emphasize task-relevant query words, such as \emph{country}, \emph{state}, \emph{lions}, \emph{score}, \emph{batter}, \emph{liquor}, and \emph{jersey}. 
In contrast, the residual components often assign larger scores to function words or generic question tokens. 
This supports our observation that the dominant query--vision interaction modes capture semantic structure useful for answering the query.

\clearpage

\end{document}